\documentclass{article} 
\usepackage{nips14submit_e,times}
\usepackage{cleveref}

\usepackage{url}
\usepackage{framed}
\usepackage{wrapfig}
\usepackage{amsthm}
\usepackage{enumitem}
\usepackage{tabularx}

\newtheorem{thm}{Theorem}

\newtheorem{assum}{Assumption}
\crefname{thm}{theorem}{theorems}
\crefname{thm}{Theorem}{Theorems}
\crefname{assum}{assumption}{assumptions}
\Crefname{assum}{Assumption}{Assumptions}

\usepackage{soul}	
\usepackage{todonotes}

\usepackage{color}
\newif\ifcomment
\commenttrue 

\ifcomment
\newcommand{\qirong}[1]{{\color{blue}{\bf\sf [Qirong: #1]}}} 
\newcommand{\qirongw}[1]{{\color{cyan}{#1}}} 
\else
\newcommand{\qirong}[1]{{\color{blue}{}}}
\newcommand{\qirongw}[1]{{\color{cyan}{}}}
\fi

\title{Distributed Machine Learning via Sufficient Factor Broadcasting}

\author{
 Pengtao Xie, Jin Kyu Kim, Yi Zhou\textsuperscript{$\star$}, Qirong Ho\textsuperscript{$\dag$}, Abhimanu  Kumar\textsuperscript{$\S$}, Yaoliang Yu, Eric Xing\\
School of Computer Science, Carnegie Mellon University;\\ \textsuperscript{$\star$}Department of EECS, Syracuse University;\\ \textsuperscript{$\dag$}Institute for Infocomm Research, A*STAR, Singapore; \textsuperscript{$\S$}Groupon \\
\{pengtaox,yaoliang,epxing\}@cs.cmu.edu;yzhou35@syr.edu;\\\{jkspruce,hoqirong,abhimanyu.kumar\}@gmail.com
}

\usepackage{amsmath}
\usepackage{amsfonts}
\usepackage{graphicx}

\newcommand{\mb}{\mathbf}

\usepackage{dsfont}
\newcommand{\RR}{\mathds{R}}

\usepackage{xspace}
\makeatletter
\DeclareRobustCommand\onedot{\futurelet\@let@token\@onedot}
\def\@onedot{\ifx\@let@token.\else.\null\fi\xspace}
\def\eg{\emph{e.g}\onedot} 
\def\ie{\emph{i.e}\onedot}

\nipsfinalcopy 

\begin{document}

\maketitle
\vspace{-0.3in}
\begin{abstract}
Matrix-parametrized models, including multiclass logistic regression and sparse coding, are used in machine learning (ML) applications ranging from computer vision to computational biology. When these models are applied to large-scale ML problems starting at millions of samples and tens of thousands of classes, their parameter matrix can grow at an unexpected rate, resulting in high parameter synchronization costs that greatly slow down distributed learning. To address this issue, we propose a Sufficient Factor Broadcasting (SFB) computation model for efficient distributed learning of a large family of matrix-parameterized models, which share the following property: the parameter update computed on each data sample is a rank-1 matrix, \ie the outer product of two ``sufficient factors" (SFs). By broadcasting the SFs among worker machines and reconstructing the update matrices
locally at each worker, SFB improves communication efficiency --- communication costs are linear in the parameter matrix's dimensions, rather than quadratic --- without affecting computational correctness. We present a theoretical convergence analysis of SFB, and empirically corroborate its efficiency on four different matrix-parametrized ML models.
\end{abstract}

\vspace{-0.2in}
\section{Introduction}
\vspace{-0.1in}

For many popular machine learning (ML) models, such as multiclass logistic regression (MLR), neural networks (NN) \cite{chilimbi2014project}, distance metric learning (DML) \cite{xing2002distance} and sparse coding \cite{olshausen1997sparse}, their parameters can be represented by a matrix $\mb{W}$. For example, in MLR, rows of $\mb{W}$ represent the classification coefficient vectors corresponding to different classes; whereas in SC rows of $\mb{W}$ correspond to the basis vectors used for reconstructing the observed data. A learning algorithm, such as stochastic gradient descent (SGD), would iteratively compute an update $\Delta \mb{W}$ from data, to be aggregated with the current version of $\mb{W}$. We call such models {\it matrix-parameterized models} (MPMs). 

Learning MPMs in large scale ML problems is challenging: ML application scales have risen dramatically, a good example being the ImageNet \cite{deng2009imagenet} compendium with millions of images grouped into tens of thousands of classes. To ensure fast running times when scaling up MPMs to such large problems, it is desirable to turn to distributed computation; however, a unique challenge to MPMs is that the parameter matrix grows rapidly with problem size,
causing straightforward parallelization strategies to perform less ideally. Consider a data-parallel algorithm, in which every worker uses a subset of the data to update the parameters ---
a common paradigm is to synchronize the full parameter matrix and update matrices amongst all workers
\cite{dean2008mapreduce,dean2012large,li2015malt,chilimbi2014project,sindhwani2012large,gopal2013distributed}.
However, this synchronization can quickly become a bottleneck: take MLR for example, in which the parameter matrix $\mb{W}$ is of size $J\times D$, where $J$ is the number of classes and $D$ is the feature dimensionality. In one application of MLR to Wikipedia~\cite{partalas2015lshtc}, $J=325$k and $D>10,000$, thus $\mb{W}$ contains several billion entries (tens of GBs of memory).
Because typical computer cluster networks can only transfer a few GBs per second at the most, inter-machine synchronization of $\mb{W}$ can dominate and bottleneck the actual algorithmic computation.
In recent years, many distributed frameworks have been developed for large scale machine learning, including Bulk Synchronous Parallel (BSP) systems such as Hadoop
\cite{dean2008mapreduce} and Spark \cite{zaharia2012resilient}, graph computation frameworks such as Pregel \cite{malewicz2010pregel}, GraphLab \cite{gonzalez2012powergraph}, and bounded-asynchronous key-value stores such as Yahoo LDA\cite{ahmed2012scalable}, DistBelief\cite{dean2012large}, Petuum-PS \cite{ho2013more}, Project Adam \cite{chilimbi2014project} and \cite{li2014scaling}.
When using these systems to learn MPMs, it is common to transmit the full parameter matrices $\mb{W}$ and/or matrix updates $\Delta \mb{W}$ between machines, usually in a server-client style \cite{dean2008mapreduce,dean2012large,sindhwani2012large,gopal2013distributed,chilimbi2014project,li2015malt}. As the matrices become larger due to increasing problem sizes, so do communication costs and synchronization delays --- hence,
reducing such costs is a key priority when using these frameworks. 


In this paper, we investigate the structure of matrix-parameterized models, in order to design efficient communication strategies that can be realized in distributed ML frameworks. We focus on models with a common property: when the parameter matrix $\mb{W}$ of these models is optimized with stochastic gradient descent (SGD) \cite{dean2012large,ho2013more,chilimbi2014project} or stochastic dual coordinate ascent (SDCA) \cite{hsieh2008dual,shalev2013stochastic,yang2013trading,jaggi2014communication,
hsieh2015comm}, the update $\bigtriangleup \mb{W}$ computed over one (or a few) data sample(s) is of low-rank, \eg it
can be written as the outer product of two vectors $\mb{u}$ and $\mb{v}$: $\bigtriangleup \mb{W}=\mb{u}\mb{v}^\top$. 
The vectors $\mb{u}$ and $\mb{v}$ are \textit{sufficient factors} (SF, meaning that they are sufficient
to reconstruct the update matrix $\bigtriangleup \mb{W}$).
A rich set of models \cite{olshausen1997sparse,lee1999learning,xing2002distance,
yuan2006model,chilimbi2014project} fall into this family: for instance, when solving an MLR problem using SGD, the stochastic gradient is $\bigtriangleup \mb{W}=\mb{u}\mb{v}^\top$, where $\mathbf{u}$ is the prediction probability vector and $\mathbf{v}$ is the feature vector. Similarly, when solving an $\ell_2$ regularized MLR problem using SDCA, the update matrix $\bigtriangleup \mb{W}$ also admits such as a structure, where $\mathbf{u}$ is the update vector of a dual variable and $\mathbf{v}$ is the feature vector. 
Other models include neural networks \cite{chilimbi2014project}, distance metric learning \cite{xing2002distance}, sparse coding \cite{olshausen1997sparse}, non-negative matrix factorization \cite{lee1999learning}, principal component analysis, and group Lasso \cite{yuan2006model}. 

Leveraging this property, we propose a computation model called Sufficient Factor Broadcasting (SFB), and evaluate its effectiveness in a peer-to-peer implementation (while noting that SFB can also be used in other distributed frameworks).
SFB efficiently learns parameter matrices using the SGD or SDCA algorithms, which are widely-used in distributed ML \cite{hsieh2008dual,dean2012large,ho2013more,shalev2013stochastic,yang2013trading,jaggi2014communication,chilimbi2014project,hsieh2015comm,li2015malt}.
The basic idea is as follows: 
since $\bigtriangleup\mb{W}$ can be exactly constructed from the sufficient factors,
rather than communicating the full (update) matrix between workers, we can instead broadcast only the sufficient factors and have workers reconstruct the updates.
SFB is thus highly communication-efficient; transmission costs are linear in the dimensions of the parameter matrix, and the resulting faster communication greatly reduces waiting time in synchronous systems (e.g. Hadoop and Spark), or improves parameter freshness in (bounded) asynchronous systems (e.g. GraphLab, Petuum-PS and \cite{li2014scaling}). SFs have been used to speed up some (but not all) network communication in deep learning \cite{chilimbi2014project}; our work differs primarily in that we always transmit SFs, never full matrices.

SFB does not impose strong requirements on the distributed system --- it can be used with synchronous~\cite{dean2008mapreduce,malewicz2010pregel,zaharia2012resilient}, asynchronous~\cite{gonzalez2012powergraph,ahmed2012scalable,dean2012large}, and bounded-asynchronous consistency models~\cite{BertsekasTsitsiklis89,ho2013more,terry2013replicated},
in order to trade off between system efficiency and algorithmic accuracy.
We provide theoretical analysis of SFB under synchronous and bounded-async consistency, and demonstrate that SFB learning of matrix-parametrized models significantly outperforms strategies that communicate the full parameter/update matrix,
on a variety of applications including distance metric learning \cite{xing2002distance}, sparse coding \cite{olshausen1997sparse} and unregularized/$\ell_2$-regularized multiclass logistic regression.
Using our own C++ implementations of each application, our experiments show that, for parameter matrices with 5-10 billion entries, replacing full-matrix communication with SFB improves convergence times by 3-4 fold. Notably, our SFB implementation of $\ell_2$-MLR is approximately 9 times faster than the Spark v1.3.1 implementation. We expect the performance benefit of SFB (versus full matrix communication) to improve with even larger matrix sizes.

\vspace{-0.1in}
\section{Sufficient Factor Property of Matrix-Parametrized Models}
\vspace{-0.1in}



The core goal of Sufficient Factor Broadcasting (SFB) is to reduce network communication costs for matrix-parametrized models; specifically, those that follow an optimization formulation
\begin{equation}
\textbf{(P)}\quad
\begin{matrix}
 \underset{\mb{W}}{\textrm{min}}& \frac{1}{N}\sum\limits_{i=1}^{N}f_i(\mb{W}\mb{a}_{i})+h(\mb{W})\\
\end{matrix}
\vspace{-0in}
\end{equation} 
where the model is parametrized by a matrix $\mathbf{W}\in \RR^{J\times D}$. The loss function $f_i(\cdot)$ is typically defined over a set of training samples $\{(\mb{a}_i,\mb{b}_i)\}_{i=1}^{N}$, with the dependence on $\mb{b}_i$ being suppressed. 
We allow $f_i(\cdot)$ to be either convex or nonconvex, smooth or nonsmooth (with subgradient everywhere); examples include
$\ell_2$ loss and multiclass logistic loss, amongst others. The regularizer $h(\mb{W})$ is assumed to admit an efficient proximal operator $\textrm{prox}_{h}(\cdot)$ \cite{beck2009fast}. For example, $h(\cdot)$ could be an indicator function of convex constraints, $\ell_1$-, $\ell_2$-, trace-norm, to name a few.
The vectors $\mb{a}_i$ and $\mb{b}_i$ can represent observed features, supervised information (e.g., class labels in classification, response values in regression), or even unobserved auxiliary information (such as sparse codes in sparse coding \cite{olshausen1997sparse}) associated with data sample $i$. The key property we exploit below ranges from the matrix-vector multiplication $\mb{W}\mb{a}_{i}$.
This optimization problem \textbf{(P)} can be used to represent a rich set of ML models \cite{olshausen1997sparse,lee1999learning,xing2002distance,
yuan2006model,chilimbi2014project}, such as the following:


\noindent\textbf{Distance metric learning (DML)} \cite{xing2002distance} improves the performance of other ML algorithms, by learning a new distance function that correctly represents similar and dissimilar pairs of data samples; this distance function is a matrix $\mb{W}$ that can have billions of parameters or more, depending on the data sample dimensionality. The vector $\mb{a}_{i}$ is the difference of the feature vectors in the $i$th data pair and $f_i(\cdot)$ can be either a quadratic function or a hinge loss function, depending on the similarity/dissimilarity label $\mb{b}_{i}$ of the data pair. In both cases, $h(\cdot)$ can be an $\ell_1$-, $\ell_2$-, trace-norm regularizer or simply $h(\cdot)=0$ (no regularization).

\noindent\textbf{Sparse coding (SC)} \cite{olshausen1997sparse} learns a dictionary of basis from data, so that the data can be re-represented sparsely (and thus efficiently) in terms of the dictionary.
In SC, $\mb{W}$ is the dictionary matrix, $\mb{a}_{i}$ are the sparse codes, $\mb{b}_{i}$ is the input feature vector and $f_i(\cdot)$ is a quadratic function \cite{olshausen1997sparse}. To prevent the entries in $\mathbf{W}$ from becoming too large, each column $\mathbf{W}_{k}$ must satisfy $\|\mathbf{W}_{k}\|_{2}\leq 1$. In this case, $h(\mb{W})$ is an indicator function which equals 0 if $\mb{W}$ satisfies the constraints and equals $\infty$ otherwise.

\vspace{-0.1in}
\subsection{Optimization via proximal SGD and SDCA}
\vspace{-0.1in}

To solve the optimization problem \textbf{(P)}, it is common to employ either (proximal) stochastic gradient descent (SGD)
\cite{dean2012large,ho2013more,chilimbi2014project,li2015malt} or stochastic dual coordinate ascent (SDCA) 
\cite{hsieh2008dual,shalev2013stochastic,yang2013trading,jaggi2014communication,hsieh2015comm}, both of which are popular and well-established parallel optimization techniques.

\noindent\textbf{Proximal SGD:} In proximal SGD, a stochastic estimate of the gradient, $\bigtriangleup\mb{W}$, is first computed over one data sample (or a mini-batch of samples), in order to update $\mb{W}$ via $\mb{W} \leftarrow \mb{W}-\eta\bigtriangleup\mb{W}$ (where $\eta$ is the learning rate). Following this, the proximal operator $\textrm{prox}_{\eta h}(\cdot)$ is applied to $\mb{W}$.
Notably, the stochastic gradient $\bigtriangleup\mb{W}$ in \textbf{(P)} can be written as the outer product of two vectors $\bigtriangleup\mb{W}=\mb{u}\mb{v}^\top$, where $\mb{u}=\frac{\partial f(\mb{W}\mb{a}_{i},\mb{b}_i)}{\partial (\mb{W}\mb{a}_{i})}$, $\mb{v}=\mb{a}_{i}$, according to the chain rule. Later, we will show that this low rank structure of $\bigtriangleup\mb{W}$ can greatly reduce inter-worker communication.

\noindent\textbf{Stochastic DCA:} SDCA
applies to problems \textbf{(P)} where $f_i(\cdot)$ is convex and $h(\cdot)$ is strongly convex 
\cite{yang2013trading} (\eg when $h(\cdot)$ contains the squared $\ell_2$ norm%
);
it solves the dual problem of \textbf{(P)}, via stochastic coordinate ascent on the dual variables. More specifically, introducing the dual matrix $\mb{U} = [\mb{u}_1, \ldots, \mb{u}_N] \in \RR^{J\times N}$ and the data matrix $\mb{A} = [\mb{a}_1, \ldots, \mb{a}_N] \in \RR^{D\times N}$, we can write the dual problem of \textbf{(P)} as
\begin{equation}
\textbf{(D)}\quad
\begin{matrix}
 \underset{\mb{U}}{\textrm{min}}& \frac{1}{N}\sum\limits_{i=1}^{N}f_i^*(-\mb{u}_{i}) + h^*(\frac{1}{N} \mb{U} \mb{A}^\top)
\end{matrix}
\vspace{-0in}
\end{equation} 
where 
$f_i^*(\cdot)$ and $h^*(\cdot)$ are the Fenchel conjugate functions of $f_i(\cdot)$ and $h(\cdot)$, respectively. The primal-dual matrices $\mb{W}$ and $\mb{U}$ are connected by\footnote{The strong convexity of $h$ is equivalent to the smoothness of the conjugate function $h^*$.} $\mb{W} = \nabla h^*(\mb{Z})$, where the auxiliary matrix $\mb{Z} := \tfrac{1}{N}\mb{U}\mb{A}^\top$.
Algorithmically, we need to update the dual matrix $\mb{U}$, the primal matrix $\mb{W}$, and the auxiliary matrix $\mb{Z}$: 
every iteration, we pick a random data sample $i$, and compute the stochastic update $\bigtriangleup \mb{u}_{i}$ by minimizing \textbf{(D)} while holding $\{\mb{u}_j\}_{j \ne i}$ fixed. The dual variable is updated via $\mb{u}_{i} \gets \mb{u}_{i} - \bigtriangleup \mb{u}_{i}$, 
the auxiliary variable via $\mb{Z} \gets \mb{Z} - \bigtriangleup\mb{u}_{i}\mb{a}^\top_i$,
and the primal variable via $\mb{W} \gets \nabla h^*(\mb{Z})$. Similar to SGD, the update of the SDCA auxiliary variable $\mb{Z}$ is also the outer product of two vectors: $\bigtriangleup\mb{u}_{i}$ and $\mb{a}_i$, which can be exploited to reduce communication cost.

\noindent\textbf{Sufficient Factor property in SGD and SDCA:}
In both SGD and SDCA, the parameter matrix update can be computed as the outer product of two vectors --- we call these sufficient factors (SFs). This property can be leveraged to improve the communication efficiency of distributed ML systems: instead of communicating parameter/update matrices among machines, we can communicate the SFs and reconstruct the update matrices locally at each machine.
Because the SFs are much smaller in size, synchronization costs can be dramatically reduced. See \Cref{sec:theory} below for a detailed analysis.

\noindent\textbf{Low-rank Extensions:}
More generally, the update matrix $\bigtriangleup \mb{W}$ may not be exactly rank-1, but still of very low rank. 
For example, when each machine uses a mini-batch of size $K$, $\bigtriangleup \mb{W}$ is of rank at most $K$; in Restricted Boltzmann Machines \cite{smolensky1986information}, the update of the weight matrix is computed from four vectors $\mb{u}_1,\mb{v}_1,\mb{u}_2,\mb{v}_2$ as $\mb{u}_1\mb{v}_1^\top-\mb{u}_2\mb{v}_2^\top$, \ie rank-2; for the BFGS algorithm \cite{bertsekas1999nonlinear}, the update of the inverse Hessian is computed from two vectors $\mb{u},\mb{v}$ as $\alpha\mb{u}\mb{u}^\top-\beta(\mb{u}\mb{v}^\top+\mb{v}\mb{u}^\top)$, \ie rank-3.
Even when the update matrix $\bigtriangleup \mb{W}$ is not genuinely low-rank, to reduce communication cost, it might still make sense to send only a certain low-rank \emph{approximation}. 
We intend to investigate these possibilities in future work.



\vspace{-0.1in}
\section{Sufficient Factor Broadcasting}
\vspace{-0.1in}


\begin{wrapfigure}{r}{0.4\textwidth}
\vspace{-0.1in}
\centering
\includegraphics[width=0.4\columnwidth]{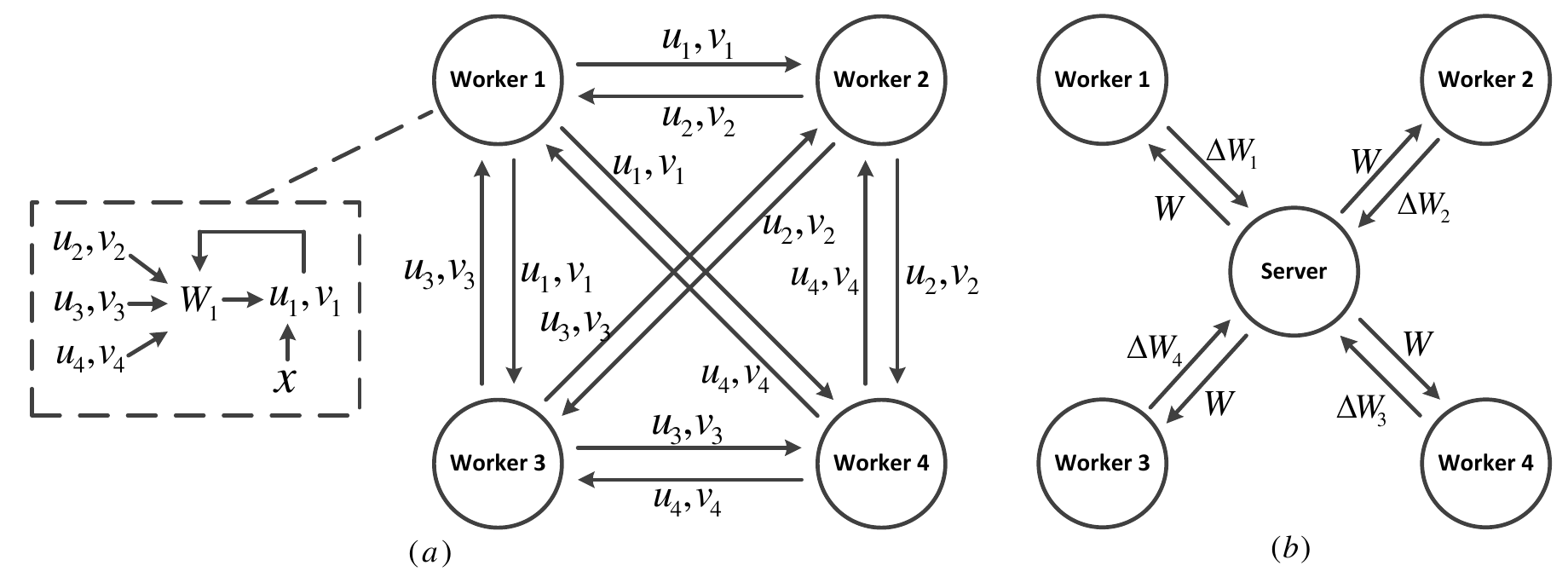}
\caption{Sufficient Factor Broadcasting (SFB).}
\label{fig:svs}\vspace{-0.15in}
\end{wrapfigure}

Leveraging the SF property of the update matrix in problems \textbf{(P)} and \textbf{(D)}, we propose a \textit{Sufficient Factor Broadcasting} (SFB) model of computation, that supports efficient (low-communication) distributed learning of the parameter matrix $\mb{W}$.
We assume a setting with $P$ workers, each of which holds a data shard and a copy of the parameter matrix\footnote{For simplicity, we assume each worker has enough memory to hold a full copy of the parameter matrix $\mb{W}$. If $\mb{W}$ is too large, one can either partition it across multiple machines \cite{dean2012large,li2014scaling,lee2014strads}, or use local disk storage (\ie out of core operation). We plan to investigate these strategies as future work.} $\mb{W}$. Stochastic updates to $\mb{W}$ are generated via proximal SGD or SDCA, and communicated between machines to ensure parameter consistency.
In proximal SGD, on every iteration, each worker $p$ computes SFs $(\mathbf{u}_{p},\mathbf{v}_{p})$, based on one data sample $\mb{x}_i = (\mb{a}_i,\mb{b}_i)$ in the worker's data shard. The worker then broadcasts $(\mathbf{u}_{p},\mathbf{v}_{p})$
to all other workers; once all $P$ workers have performed their broadcast (and have thus received all SFs),
they re-construct the $P$ update matrices (one per data sample) from the $P$ SFs, and 
apply them to update their local copy of $\mb{W}$. Finally, each worker applies
the proximal operator $\textrm{prox}_{h}(\cdot)$.
When using SDCA, the above procedure is instead used to broadcast SFs for the auxiliary matrix $\mb{Z}$, which is then used to obtain the primal matrix $\mb{W} = \nabla h^*(\mb{Z})$.
\Cref{fig:svs} illustrates SFB operation: 4 workers compute their respective SFs $(\mb{u}_1,\mb{v}_1)$, $\dots$, $(\mb{u}_4,\mb{v}_4)$, which are then broadcast to the other 3 workers.
Each worker $p$ uses all 4 SFs $(\mb{u}_1,\mb{v}_1), \dots, (\mb{u}_4,\mb{v}_4)$
to exactly reconstruct the update matrices $\bigtriangleup \mb{W}_{p} = \mb{u}_{p} \mb{v}_{p}^\top$, and update their local copy of the parameter matrix: $\mb{W}_{p} \leftarrow \mb{W}_{p} - \sum_{q=1}^4 \mb{u}_{q} \mb{v}_{q}^\top$.  While the above description reflects synchronous execution, asynchronous and bounded-asynchronous extensions are also possible (Section \ref{sec:sfbcaster}).

\noindent\textbf{SFB vs client-server architectures:}
The SFB peer-to-peer topology can be contrasted with a ``full-matrix" client-server architecture for parameter synchronization, e.g. as used by Project Adam \cite{chilimbi2014project} to learn neural networks: there, a centralized server maintains the global parameter matrix, and each client keeps a local copy. Clients compute sufficient factors and send them to the server, which uses the SFs to update the global parameter matrix; the server then sends the full, updated parameter matrix back to clients. Although client-to-server costs are reduced (by sending SFs), server-to-client costs are still expensive because full parameter matrices need to be sent. In contrast, the peer-to-peer SFB topology never sends full matrices; only SFs are sent over the network.
We also note that under SFB, the update matrices are reconstructed at each of the $P$ machines, rather than once at a central server (for full-matrix architectures). 
Our experiments show that the time taken for update reconstruction is empirically negligible compared to communication and SF computation.

\noindent\textbf{Mini-batch proximal SGD/SDCA:}
SFB can also be used in mini-batch proximal SGD/SDCA; every iteration, each worker samples a mini-batch of $K$ data points, and computes $K$ pairs of sufficient factors $\{(\mb{u}_{i}, \mb{v}_{i})\}_{i=1}^{K}$. These $K$ pairs are broadcast to all other workers, which reconstruct the originating worker's update matrix as $\bigtriangleup\mb{W}=\frac{1}{K}\sum_{i=1}^{K}\mb{u}_{i}\mb{v}_{i}^{\mathsf{T}}$.



\vspace{-0.1in}
\section{Sufficient Factor Broadcaster: An Implementation}
\vspace{-0.1in}
\label{sec:sfbcaster}
In this section, we present Sufficient Factor Broadcaster (SFBcaster) --- an implementation of SFB --- including consistency models, programing interface and implementation details.
We stress that SFB does not require a special system; it can be implemented on top of existing distributed frameworks, using any suitable communication topology --- such as star\footnote{For example, each worker sends the SFs to a hub machine, which re-broadcasts them to all other workers.}, ring, tree, fully-connected and Halton-sequence~\cite{li2015malt}. As future work, we intend to investigate the effect of these topologies on performance.

\vspace{-0.1in}
\subsection{Flexible Consistency Models}
\label{sec:consist}
\vspace{-0.1in}
Our SFBcaster implementation supports three consistency models: Bulk Synchronous Parallel (BSP-SFB), Asynchronous Parallel (ASP-SFB), and Stale Synchronous Parallel (SSP-SFB), and we provide theoretical convergence guarantees for BSP-SFB and SSP-SFB in the next section.
\begin{wrapfigure}{r}{0.45\textwidth}
\begin{framed}
\textcolor{blue}{sfb\_app} mlr ( \textcolor{blue}{int} J, \textcolor{blue}{int} D, \textcolor{blue}{int} staleness )\\*
\textcolor{orange}{//SF computation function}\\*
function compute\_sv ( \textcolor{blue}{sfb\_app} mlr ):\\*
 \hspace*{0.2cm} while ( ! converged ):\\
\hspace*{0.4cm} $X$ = sample\_minibatch ()\\
\hspace*{0.4cm} foreach $\mb{x}_i$ in $X$:\\
 \hspace*{0.6cm} \textcolor{orange}{//sufficient factor $\mb{u}_i$}\\
\hspace*{0.6cm} pred = predict ( mlr.para\_mat, $\mb{x}_i$ ) \\            
\hspace*{0.6cm} mlr.sv\_list[i].write\_u ( pred ) \\
  \hspace*{0.6cm} \textcolor{orange}{//sufficient factor $\mb{v}_i$}\\
  \hspace*{0.6cm} mlr.sv\_list[i].write\_v ( $\mb{x}_i$ ) \\
  \hspace*{0.4cm} commit() \\
\vspace{-0.1in}
\end{framed}
\vspace{-0.2in}
\caption{Multiclass LR pseudocode.}
\label{fig:mlr}
\vspace{-0.1in}
\end{wrapfigure}
\begin{wrapfigure}{r}{0.55\textwidth}
\vspace{-0.2in}
\begin{framed}
\textcolor{blue}{sfb\_app} sc ( \textcolor{blue}{int} D, \textcolor{blue}{int} J , \textcolor{blue}{int} staleness)\\*
\textcolor{orange}{//SF computation function}\\*
 function compute\_sf ( \textcolor{blue}{sfb\_app} sc ):\\*
 \hspace*{0.2cm} while ( ! converge ): \\
\hspace*{0.4cm} $X$=sample\_minibatch ()\\
\hspace*{0.4cm} foreach $\mb{x}_i$ in $X$:\\
 \hspace*{0.6cm} \textcolor{orange}{//compute sparse code}\\
\hspace*{0.6cm} a = compute\_sparse\_code ( sc.para\_mat, $\mb{x}_i$ )   \\   
\hspace*{0.6cm} \textcolor{orange}{//sufficient factor $\mb{u}_i$}\\         
\hspace*{0.6cm} sc.sf\_list[i].write\_u (  sc.para\_mat * a-$\mb{x}_i$ )\\
  \hspace*{0.6cm} \textcolor{orange}{//sufficient factor $\mb{v}_i$}\\
  \hspace*{0.6cm} sc.sf\_list[i].write\_v ( a )\\
  \hspace*{0.4cm} commit()\\
\textcolor{orange}{//Proximal operator function}\\*
 function prox ( \textcolor{blue}{sfb\_app} sc ):\\*
\hspace*{0.2cm} foreach column $\mb{b}_i$ in sc.para\_mat:\\
  \hspace*{0.4cm} if $\|\mb{b}_i\|_2>1$:\\
  \hspace*{0.6cm} $\mb{b}_i=\frac{\mb{b}_i}{\|\mb{b}_i\|_2}$\\
\vspace{-0.1in}
\end{framed}
\vspace{-0.1in}
\label{fig:sc}
\caption{Sample code of sparse coding in SFB}
\vspace{-0.2in}
\end{wrapfigure}
\noindent\textbf{BSP-SFB:}
Under BSP \cite{dean2008mapreduce,malewicz2010pregel,zaharia2012resilient},
an end-of-iteration global barrier ensures all workers have completed their work, and synchronized their parameter copies, before proceeding to the next iteration. BSP is a strong consistency model, that guarantees the same computational outcome (and thus algorithm convergence) each time.

\noindent\textbf{ASP-SFB:}
BSP can be sensitive to stragglers (slow workers) \cite{ho2013more,terry2013replicated}, 
limiting the distributed system to the speed of the slowest worker.

The Asynchronous Parallel (ASP) \cite{gonzalez2012powergraph,ahmed2012scalable,dean2012large} communication model addresses this issue, by allowing workers to proceed without waiting for others.
ASP is efficient in terms of iteration throughput, but carries the risk that worker parameter copies can end up greatly out of synchronization, which can lead to algorithm divergence \cite{ho2013more}. 

\noindent\textbf{SSP-SFB:}
Stale Synchronous Parallel (SSP) \cite{BertsekasTsitsiklis89,ho2013more,terry2013replicated} is a bounded-asynchronous consistency model that serves as a middle ground between BSP and ASP; it allows workers to advance at different rates, provided that the difference in iteration number between the slowest and fastest workers is no more than a user-provided staleness $s$. SSP alleviates the straggler issue while guaranteeing algorithm convergence \cite{ho2013more,terry2013replicated}. Under SSP-SFB, each worker $p$ tracks the number of SF pairs computed by itself, $t_p$, versus the number $\tau_p^{q}(t_p)$ of SF pairs received from each worker $q$. If there exists a worker $q$ such that $t_p-\tau_p^{q}(t_p)> s$ (\ie some worker $q$ is likely more than $s$ iterations behind worker $p$), then worker $p$ pauses until $q$ is no longer $s$ iterations or more behind.
When $s=0$, SSP-SFB reduces to BSP-SFB \cite{dean2008mapreduce,zaharia2012resilient}, and when $s=\infty$, SSP-SFB becomes ASP-SFB.

\vspace{-0.1in}
\subsection{Programming Interface}
\vspace{-0.1in}

The SFBcaster programming interface is simple; users need to provide a SF computation function to specify how to compute the sufficient factors. 
To send out SF pairs $(\mb{u},\mb{v})$, the user adds them to a buffer object $sv\_list$, via:
$write\_u(vec\_u)$, $write\_v(vec\_v)$, which set $i$-th SF $\mb{u}$ or $\mb{v}$ to $vec\_u$ or $vec\_v$.
All SF pairs 
are sent out at the end of an iteration, which is signaled by
$commit()$.
Finally, in order to choose between BSP, ASP and SSP consistency, users simply set $staleness$ to an appropriate value (0 for BSP, $\infty$ for ASP, all other values for SSP).
SFBcaster automatically updates each worker's local parameter matrix using all SF pairs --- including both locally computed SF pairs added to $sv\_list$, as well as SF pairs received from other workers.

Figure \ref{fig:mlr} shows SFBcaster pseudocode for multiclass logistic regression. For proximal SGD/SDCA algorithms, SFBcaster requires users to write an additional function, $prox(mat)$, which applies the proximal operator $\textrm{prox}_{h}(\cdot)$ (or the SDCA dual operator $h^*(\cdot)$) to the parameter matrix $mat$. Figure 3 shows the sample code of implementing sparse coding in SFB.  $D$ is the feature dimensionality of data and $J$ is the dictionary size. Users need to write a SF computation function to specify how to compute the sufficient factors: for each data sample $\mb{x}_{i}$, we first compute its sparse code $\mb{a}$ based on the dictionary $\mb{B}$ stored in the parameter matrix $\textrm{sc.para\_mat}$. Given $\mb{a}$, the sufficient factor $\mb{u}$ can be computed as $\mb{B}\mb{a}-\mb{x}_{i}$ and the sufficient factor $\mb{v}$ is simply $\mb{a}$. In addition, users need to provide a proximal operator function to specify how to project $\mb{B}$ to the $\ell_2$ ball constraint set.
\vspace{-0.1in} 
\section{Implementation Details}
\vspace{-0.1in} 
\begin{figure}
\begin{center}
\includegraphics[width=0.4\columnwidth]{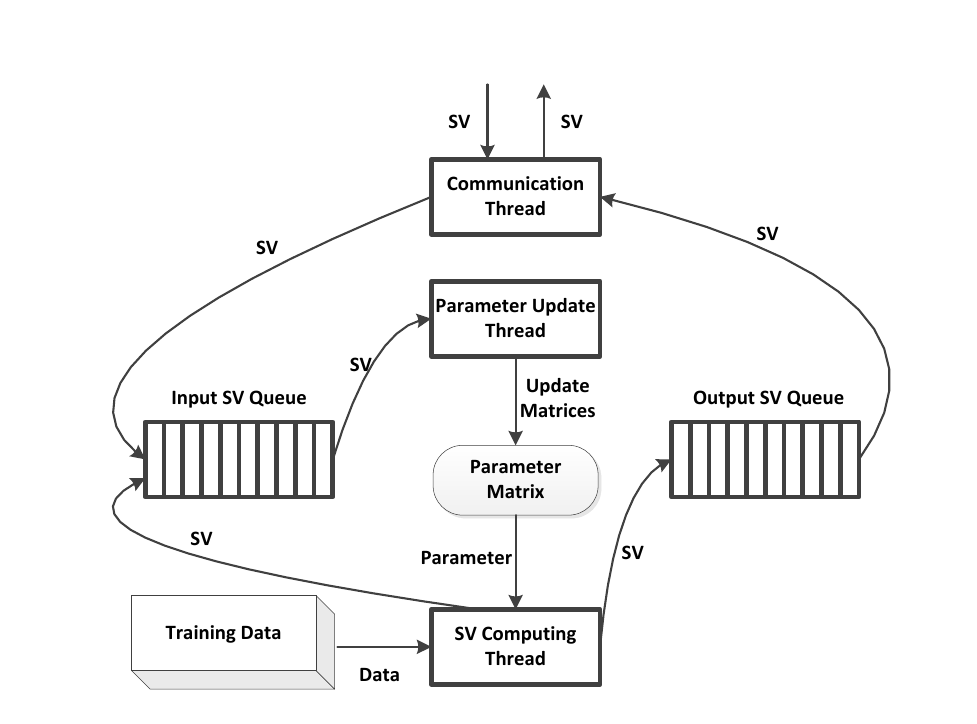}
\caption{Implementation details on each worker in SFBcaster.}
\label{fig:imple}
\end{center}
\end{figure}
Figure \ref{fig:imple} shows the implementation details on each worker in SFBcaster.
Each worker maintains three threads: SF computing thread, parameter update thread and communication thread. Each worker holds a local copy of the parameter matrix and a partition of the training data. It also maintains an input SF queue which stores the sufficient factors computed locally and received remotely and an output SF queue which stores SFs to be sent to other workers. In each iteration, the SF computing thread checks the consistency policy detailed in Section 4 in the main paper. If permitted, this thread randomly chooses a minibatch of samples from the training data, computes the SFs and pushes them to the input and output SF queue. The parameter update thread fetches SFs from the input SF queue and uses them to update the parameter matrix. In proximal-SGD/SDCA, the proximal/dual operator function (provided by the user) is automatically called by this thread as a function pointer. 
The communication thread receives SFs from other workers and pushes them into the input SF queue and broadcasts SFs in the output SF queue to other workers. One worker is in charge of measuring the objective value. Once the algorithm converges, this worker notifies all other workers to terminate the job. We implemented SFBcaster in C++. OpenMPI was used for communication between workers and OpenMP was used for multicore parallelization within each machine.

The decentralized architecture of SFBcaster makes it robust to machine failures. If one worker fails, the rest of workers can continue to compute and broadcast the sufficient factors among themselves. In addition, SFBcaster possesses high elasticity \cite{li2014scaling}: new workers can be added and existing workers can be taken offline, without restarting the running framework.
A thorough study of fault tolerance and elasticity will be left for future work.

\vspace{-0.1in}
\section{Cost Analysis and Theory}
\label{sec:theory}
\vspace{-0.1in}

\begin{figure}
\small
\begin{tabularx}{\textwidth}{ X|X|X|X }
\hline
{\bf Computational Model} & Total comms, per iter & $\mb{W}$ storage per machine & $\mb{W}$ update time, per iter \\
\hline\hline
SFB (peer-to-peer)  &  $O(P^2K(J+D))$ & $O(JD)$ & $O(P^2KJD)$ \\
\hline
FMS (client-server \cite{chilimbi2014project}) & $O(PJD)$ & $O(JD)$ & $O(PJD)$ at server, $O(PKJD)$ at clients \\
\hline
\end{tabularx}
\vspace{-0.15in}
\caption{Cost of using SFB versus FMS, where $K$ is minibatch size, $J,D$ are dimensions of $\mb{W}$, and $P$ is the number of workers.
}
\label{tab:comms_cost}
\vspace{-0.2in}
\end{figure}

We now examine the costs and convergence behavior of SFB under synchronous and bounded-async (e.g. SSP~\cite{BertsekasTsitsiklis89,ho2013more,dai2015high}) consistency, and show that SFB can be preferable to full-matrix synchronization/communication schemes.

\noindent\textbf{Cost Analysis}:
Table \ref{tab:comms_cost} compares the communications, space and time (to apply updates to $\mb{W}$) costs of peer-to-peer SFB, against full matrix synchronization (FMS) under a client-server architecture \cite{chilimbi2014project}. For SFB, in each minibatch, every worker broadcasts $K$ SF pairs $(\mb{u},\mb{v})$ to $P-1$ other workers, i.e. $O(P^2K(J+D))$ values are sent per iteration ---
linear in matrix dimensions $J,D$, and quadratic in $P$. Because SF pairs cannot be aggregated before transmission, the cost has a dependency on $K$.
In contrast, the communication cost in FMS is $O(PJD)$, linear in $P$, quadratic in matrix dimensions, and independent of $K$. For both SFB and FMS, the cost of storing $\mb{W}$ is $O(JD)$ on every machine. As for the time taken to update $\mb{W}$ per iteration, FMS costs $O(PJD)$ at the server (to aggregate $P$ client update matrices) and $O(PKJD)$ at the $P$ clients (to aggregate $K$ updates into one update matrix for the server). By comparison, SFB bears a cost of $O(P^2KJD)$ due to the additional overhead of reconstructing each update matrix $P$ times.

Compared with FMS, SFB achieves communication savings by paying an extra computation cost. 
In a number of practical scenarios, such a tradeoff is worthwhile. Consider large problem scales where $\min(J,D)\ge 10000$, and moderate minibatch sizes $1\le K\le 1000$ (as studied in this paper); when using a moderate number of machines (around 10-100), the $O(KP^2(J+D))$ communications cost of SFB is lower than the $O(PJD)$ cost for FMS, and the relative benefit of SFB improves as the dimensions $J,D$ of $\mb{W}$ grow. As for the time needed to apply updates to $\mb{W}$, it turns out that the additional cost of reconstructing each update matrix $P$ times in SFB is negligible in practice --- we have observed in our experiments that the time spent computing SFs, as well as communicating SFs over the network, greatly dominates the cost of reconstructing update matrices using SFs. Overall, the communication savings dominate the added computational overhead, which we validated in experiments (Section \ref{sec:exp}).

In order to make SFB applicable to data centers with 1000s of machines \cite{dean2012large,li2014scaling}, we note that SFB costs can be further reduced via efficient broadcast strategies: e.g. in the Halton-sequence strategy \cite{li2015malt}, each machine connects with and broadcasts messages to $Q = O(\log(P))$ machines, rather than all $P$ machines. We leave the issue of broadcast strategies as future study.

\noindent\textbf{Convergence Analysis:}
We study the convergence of SF minibatch SGD (with extensions to proximal-SGD and SDCA being a topic for future study).
We wish to solve the optimization problem
$\min_{\mb{W}}~ \sum_{m=1}^{M} f_m(\mb{W})$,
where $M$ is the number of training data minibatches, and $f_m$ corresponds to the loss function on the $m$-th minibatch. Assume the training data minibatches $\{1,...,M\}$ are divided into $P$ disjoint subsets $\{S_1,...,S_P\}$ with $|S_p|$ denoting the number of minibatches in $S_p$. Denote $F =  \sum_{m=1}^{M} f_m$ as the total loss, and for $p=1,\ldots, P$, $F_p := \sum_{j\in S_p} f_j$ is the loss on $S_p$ (residing on the $p$-th machine). 

Consider a distributed system with $P$ machines. Each machine $p$ keeps a local variable $\mathbf{W}_p$ and the training data in $S_p$. At each iteration, machine $p$ draws one minibatch $I_p$ uniformly at random from partition $S_p$, and computes the partial gradient $\sum_{j\in I_p}\nabla f_j (\mathbf{W}_p)$. 
Each machine updates its local variable by accumulating partial  updates from all machines. Denote $\eta_c$ as the learning rate at $c$-th iteration on every machine. The partial update generated by machine $p$ at its $c$-th iteration is 
denoted as $U_p(\mb{W}_p^c, I_p^c) = -\eta_c |S_p| \sum_{j\in I_p^c}\nabla f_{j}(\mathbf{W}_p^c)$. Note that $I_p^c$ is random and the factor $|S_p|$ is to restore unbiasedness in expectation.
Then the local update rule of machine $p$ is 
\begin{flalign}
\label{eq:local}
\textstyle\mathbf{W}_p^c = \mathbf{W^0} + \sum_{q=1}^{P}\sum_{t=0}^{\tau_{p}^q(c)}  U_q(\mb{W}_q^t, I_q^t), \quad 0\le (c-1)-\tau_{p}^q(c) \le s,
\end{flalign}
where $\mb{W}^0$ is the common initializer for all $P$ machines, and $\tau_{p}^q(c)$ is the number of iterations machine $q$ has transmitted to machine $p$ when machine $p$ conducts its $c$-th iteration. Clearly, $\tau_p^p(c) = c$. Note that we also require $\tau_p^q(c) \leq c-1$, \ie, machine $p$ will not use any partial updates of machine $q$ that are too fast forward. This is to avoid correlation in the theoretical analysis.
Hence, machine $p$ (at its $c$-th iteration) accumulates updates generated by machine $q$ up to iteration $\tau_{p}^q(c)$, which is restricted to be at most $s$ iterations behind.
Such bounded-asynchronous communication addresses the slow-worker problem caused by bulk synchronous execution, while ensuring that the updates accumulated by each machine are not too outdated. 
The following standard assumptions are needed for our theoretical analysis:
\begin{assum}\label{ass:model_3}
(1) For all $j$, $f_j$ is continuously differentiable and $F$ is bounded from below; (2) $\nabla F$, $\nabla F_{p}$ are Lipschitz continuous with constants $L_F$ and $L_p$, respectively, and let $L = \sum_{p=1}^{P} L_p$; (3) There exists $B, \sigma^2$ such that for all $p$ and $c$, we have (almost surely) $\|\mathbf{W}_p^c\|\le B$ and $\mathbb{E}\| ~|S_p|\sum_{j\in I_p}\nabla f_{j}(\mathbf{W}) - \nabla F_{p}(\mathbf{W}) ~\|_2^2\le \sigma^2$.
\end{assum}

\begin{figure}[t]
\begin{center}
\includegraphics[width=0.24\columnwidth]{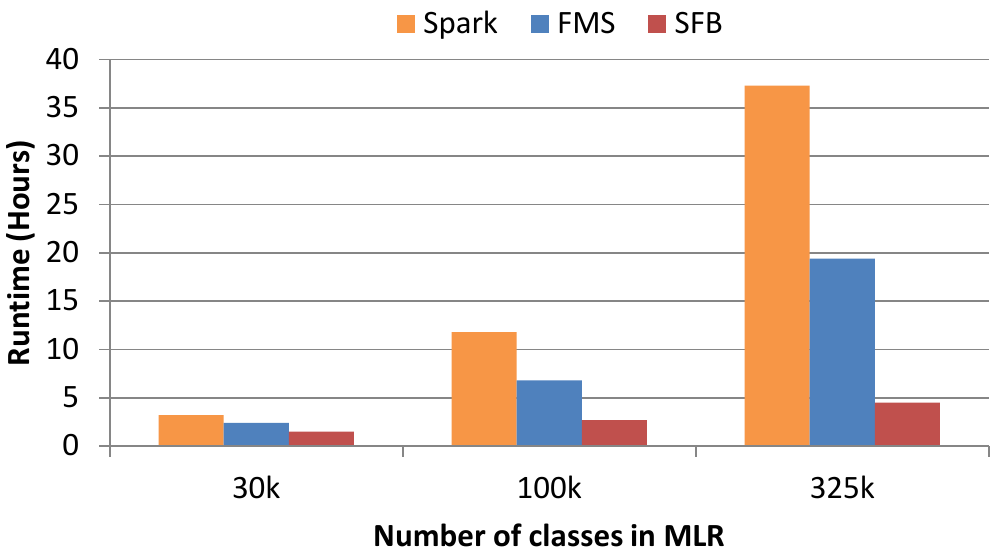}
\includegraphics[width=0.24\columnwidth]{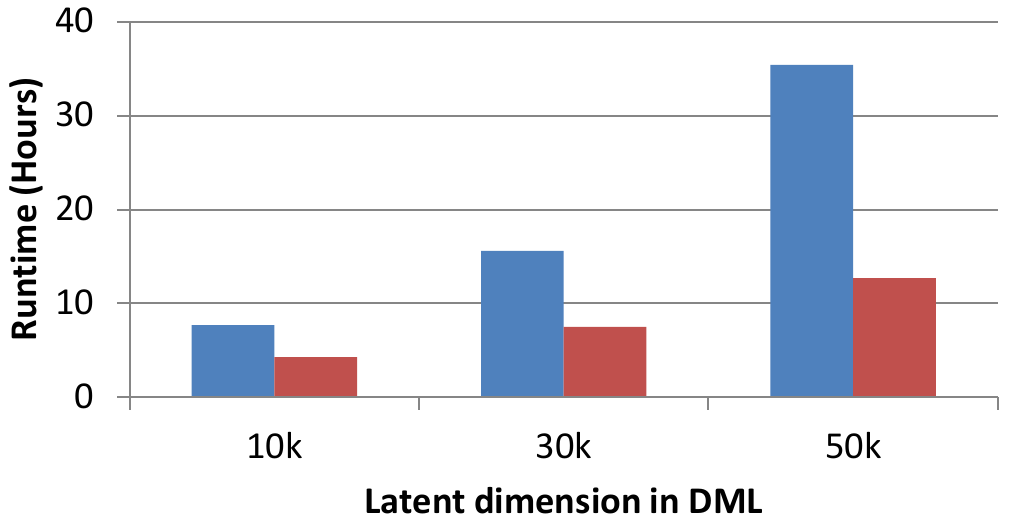}
\includegraphics[width=0.24\columnwidth]{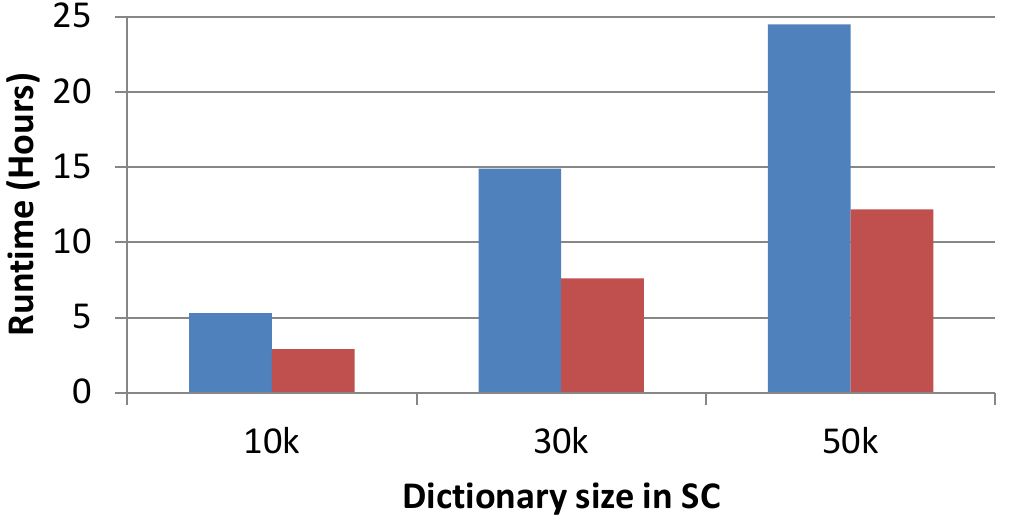}
\includegraphics[width=0.24\columnwidth]{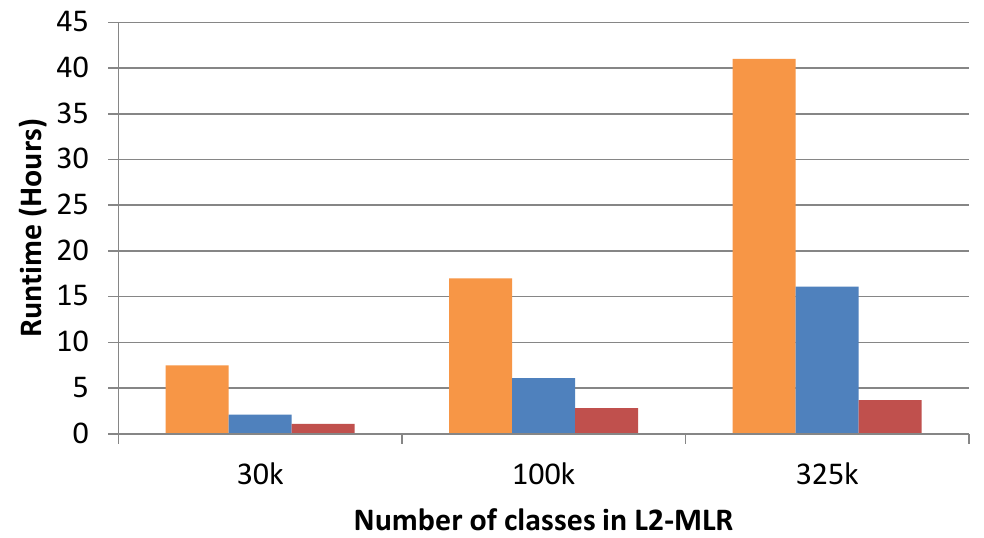}
\vspace{-0.1in}
\caption{Convergence time versus model size for MLR, DML, SC, L2-MLR (left to right), under BSP.}
\label{fig:exp_runtime}
\end{center}
\vspace{-0.1in}
\end{figure}

\begin{figure}[t]
\begin{center}
\includegraphics[width=0.24\columnwidth]{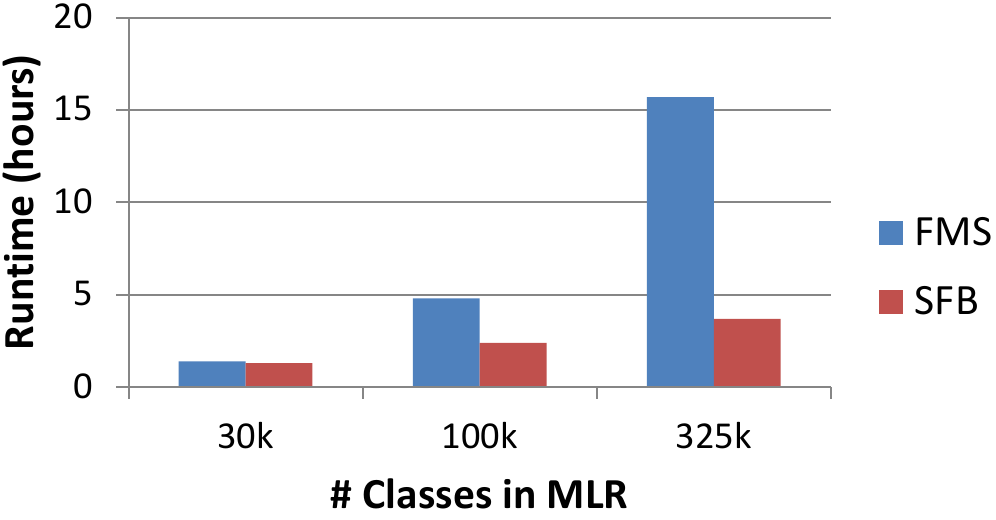}
\includegraphics[width=0.24\columnwidth]{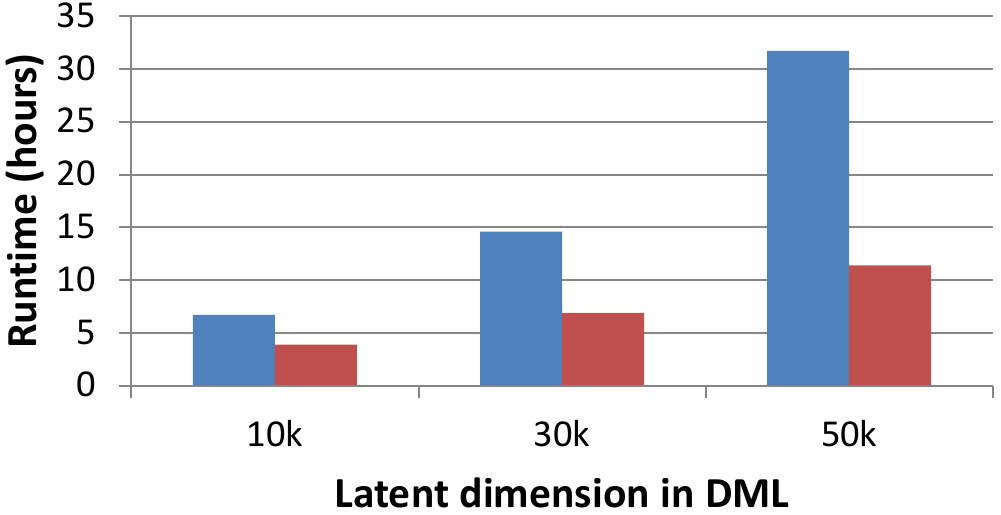}
\includegraphics[width=0.24\columnwidth]{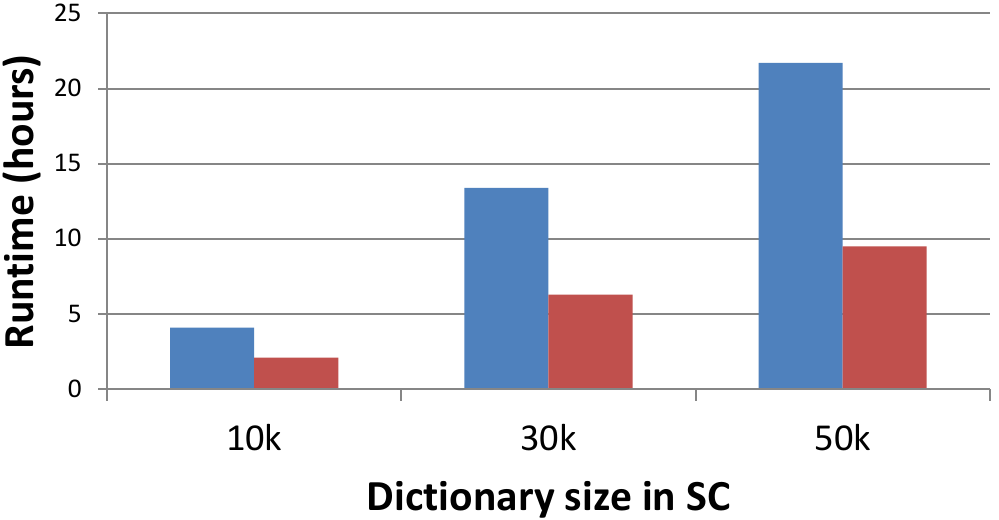}
\includegraphics[width=0.24\columnwidth]{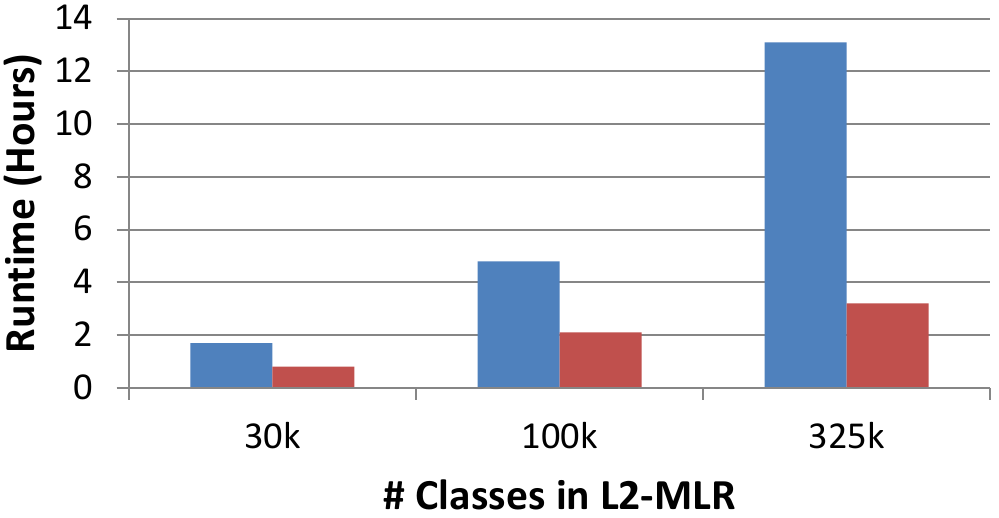}
\vspace{-0.1in}
\caption{Convergence time versus model size for MLR, DML, SC, L2-MLR (left to right), under SSP with staleness=20.}
\label{fig:exp_runtime_ssp}
\end{center}
\vspace{-0.1in}
\end{figure}

\begin{figure}[t]
\begin{center}
\includegraphics[width=0.3\columnwidth]{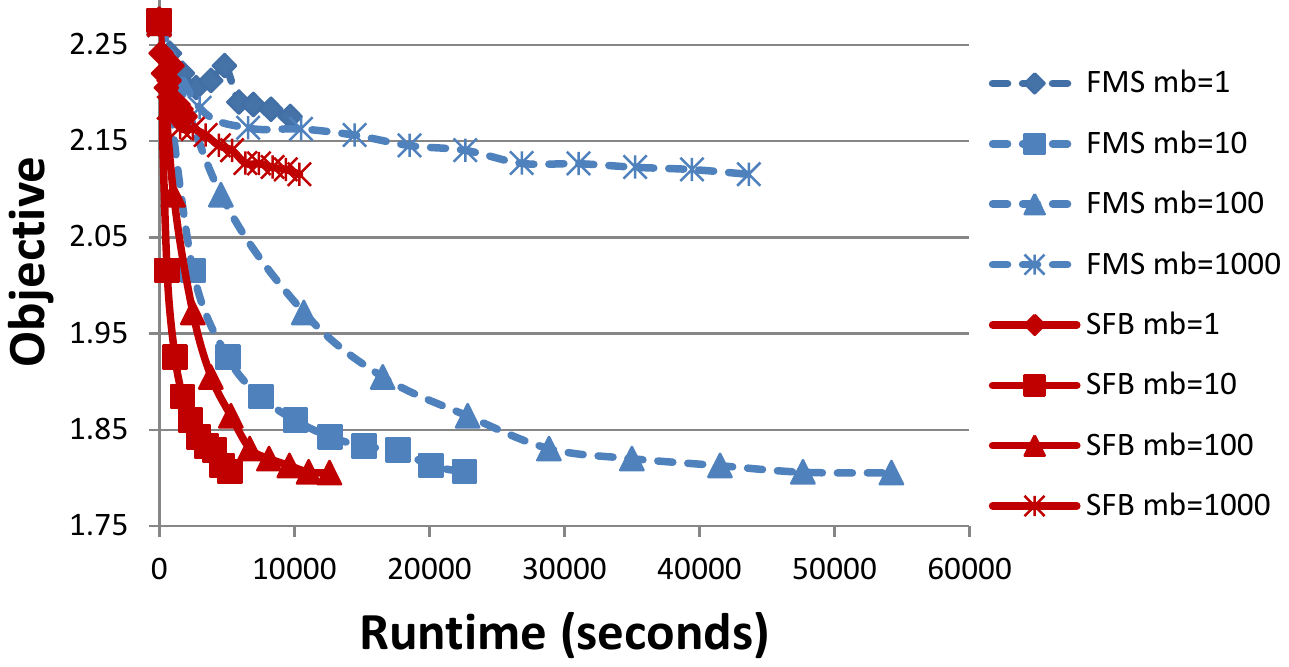}
\includegraphics[width=0.3\columnwidth]{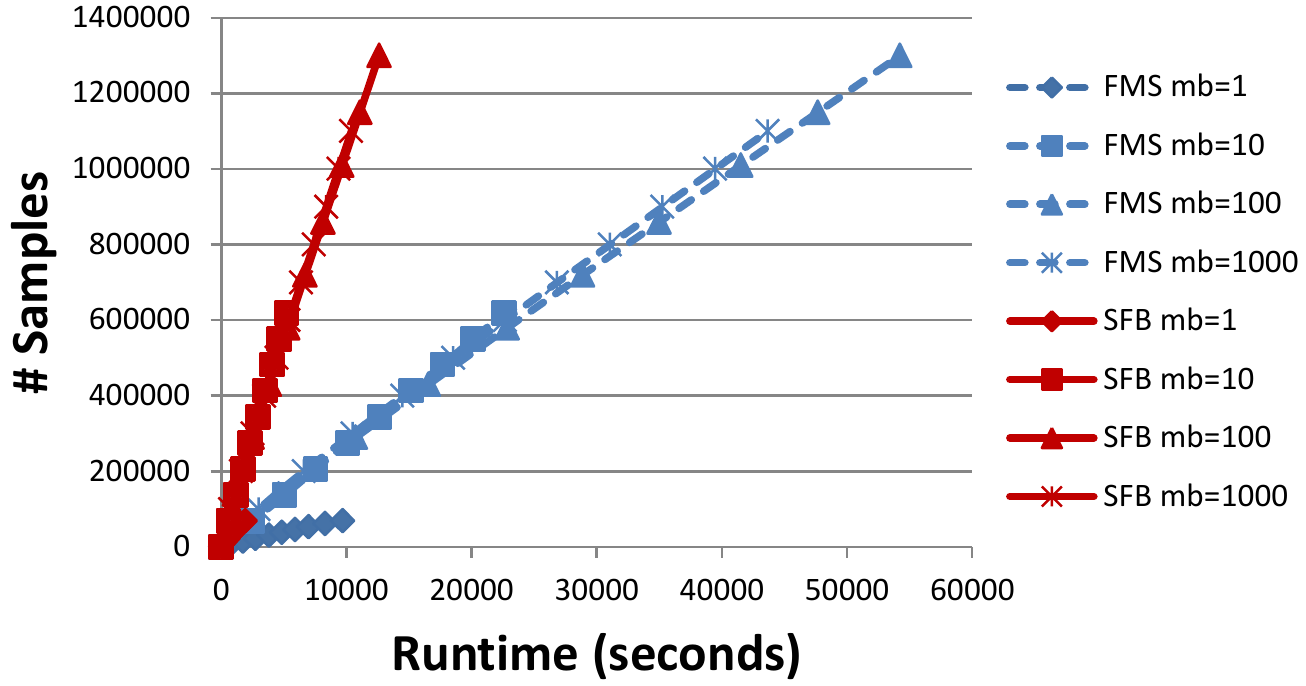}
\includegraphics[width=0.3\columnwidth]{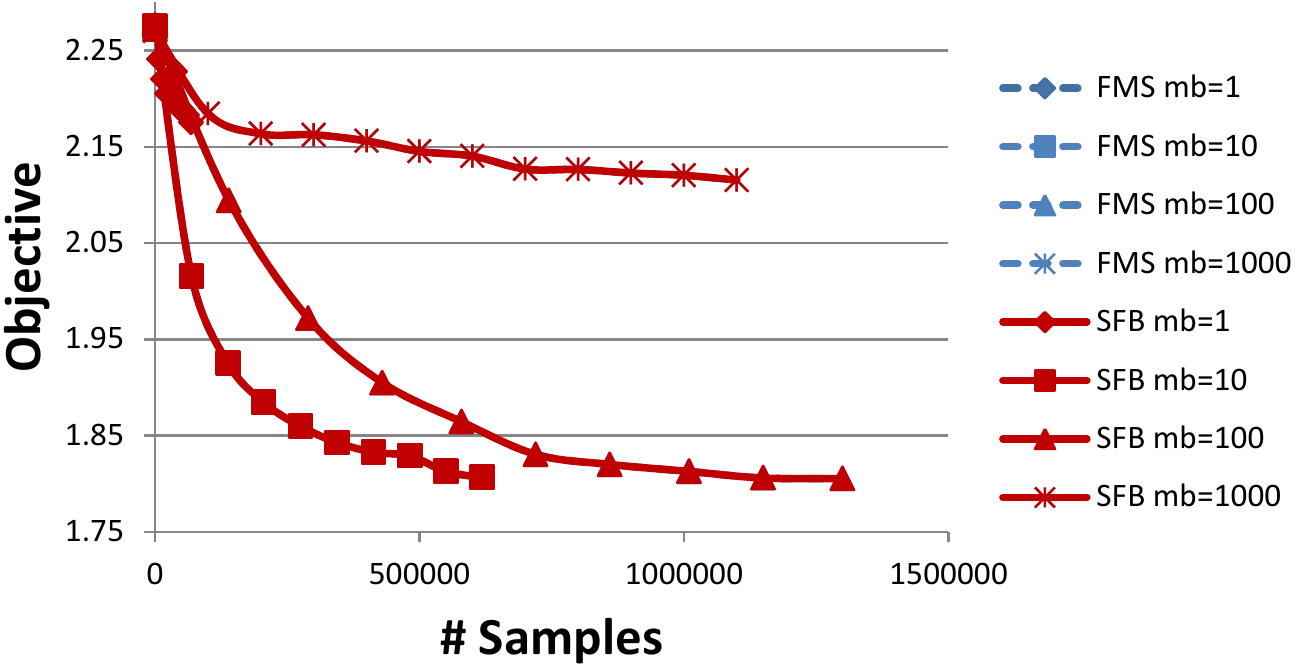}
\vspace{-0.1in}
\caption{MLR objective vs runtime (left), samples vs runtime (middle),  objective vs samples (right).}
\label{fig:iter_qtt_qlt}
\end{center}
\vspace{-0.1in}
\end{figure}

Our analysis is based on the following auxiliary update
\begin{flalign}
\vspace{-0.1in}
\textstyle\mathbf{W}^c &= \textstyle\mathbf{W}^0 + \sum_{q=1}^{P}\sum_{t=0}^{c-1} U_q(\mb{W}_q^t, I_q^t),
\vspace{-0.1in}
\end{flalign}
Compare to the local update \eqref{eq:local} on machine $p$, essentially this auxiliary update accumulates all $c-1$ updates generated by all machines, instead of the $\tau_p^q(c)$ updates that machine $p$ has access to.
We show that all local machine parameter sequences are asymptotically consistent with this auxiliary sequence:

\begin{thm}\label{thm:model_2}
Let $\{\mathbf{W}_p^c\}$, $p=1, \ldots, P$, and $\{\mathbf{W}^c\}$ be the local sequences and the auxiliary sequence generated by SFB for problem $\textbf{(P)}$ (with $h \equiv 0$), respectively.
Under \Cref{ass:model_3} and set the learning rate $\eta_c^{-1} = \frac{L_F}{2}+2sL +\sqrt{c}$, then we have
\begin{itemize}[leftmargin=*,topsep=0pt,noitemsep]
\item $\liminf\limits_{c\to\infty} \mathbb{E}\|\nabla F(\mathbf{W}^c)\|=0$, hence there exists a subsequence of $\nabla F(\mb{W}^c)$ that almost surely vanishes; 
\item $\lim\limits_{c\to\infty} \max_p \|\mathbf{W}^c - \mathbf{W}_p^c\| = 0$, \ie the maximal disagreement between all local sequences and the auxiliary sequence converges to 0 (almost surely);
\item There exists a common subsequence of $\{\mb{W}_p^c\}$ and $\{\mb{W}^c\}$ that converges almost surely to a stationary point of $F$, with the rate $\underset{c\le C}{\mathrm{min}}~\mathbb{E}\|\sum_{p=1}^{P}\nabla F_{p}(\mathbf{W}_p^c)\|_2^2 \le O\left(\frac{(L+L_F)\sigma^2 P s\log C}{\sqrt{C}}\right)$. 

\vspace{-0.1in}
\end{itemize}
\end{thm}
Intuitively, Theorem~\ref{thm:model_2} says that, given a properly-chosen learning rate, all local worker parameters $\{\mb{W}^c_p\}$ eventually converge to stationary points (i.e. local minima) of the objective function $F$, despite the fact that SF transmission can be delayed by up to $s$ iterations. Thus, SFB learning is robust even under bounded-asynchronous communication (such as SSP). 
Our analysis differs from \cite{BertsekasTsitsiklis89} in two ways:  (1) \cite{BertsekasTsitsiklis89} explicitly maintains a consensus model which would require transmitting the parameter matrix among worker machines --- a communication bottleneck that we were able to avoid;
(2) we allow subsampling in each worker machine. Accordingly, our theoretical guarantee is probabilistic, instead of the deterministic one in \cite{BertsekasTsitsiklis89}. 
Note that our analysis also covers worker parameters $\{\mb{W}^c_p\}$, which improves upon analyses that only show convergence of a central server parameter~\cite{ho2013more,dai2015high,agarwal2011distributed}.


\vspace{-0.1in}
\section{Experiments}
\label{sec:exp}
\vspace{-0.1in}

We demonstrate how four popular models can be efficiently learnt using SFB: (1) multiclass logistic regression (MLR) and distance metric learning (DML)\footnote{For DML, we use the parametrization proposed in \cite{weinberger2005distance}, which learns a linear projection matrix $\mb{L}\in \mathrm{R}^{d\times k}$, where $d$ is the feature dimension and $k$ is the latent dimension.} based on SGD; (2) sparse coding (SC) based on proximal SGD; (3) $\ell_2$ regularized multiclass logistic regression (L2-MLR) based on SDCA. For baselines, we compare with (a) Spark \cite{zaharia2012resilient} for MLR and L2-MLR, and (b) full matrix synchronization (FMS) implemented on open-source parameter servers \cite{ho2013more,li2014scaling} for all four models. In FMS, workers send update matrices to the central server, which then sends up-to-date parameter matrices to workers\footnote{This has the same communication complexity as \cite{chilimbi2014project}, which sends SFs from clients to servers, but sends full matrices from servers to clients (which dominates the overall cost).}. 
Due to data sparsity, both the update matrices and sufficient factors are sparse; we use this fact to reduce communication and computation costs.
Our experiments used a 12-machine cluster; each machine has 64 2.1GHz AMD cores, 128G memory, and a 10Gbps network interface.
\vspace{-0.1in}
\paragraph{Datasets and Experimental Setup}

\begin{figure}[t]
\begin{center}
\includegraphics[width=0.24\columnwidth]{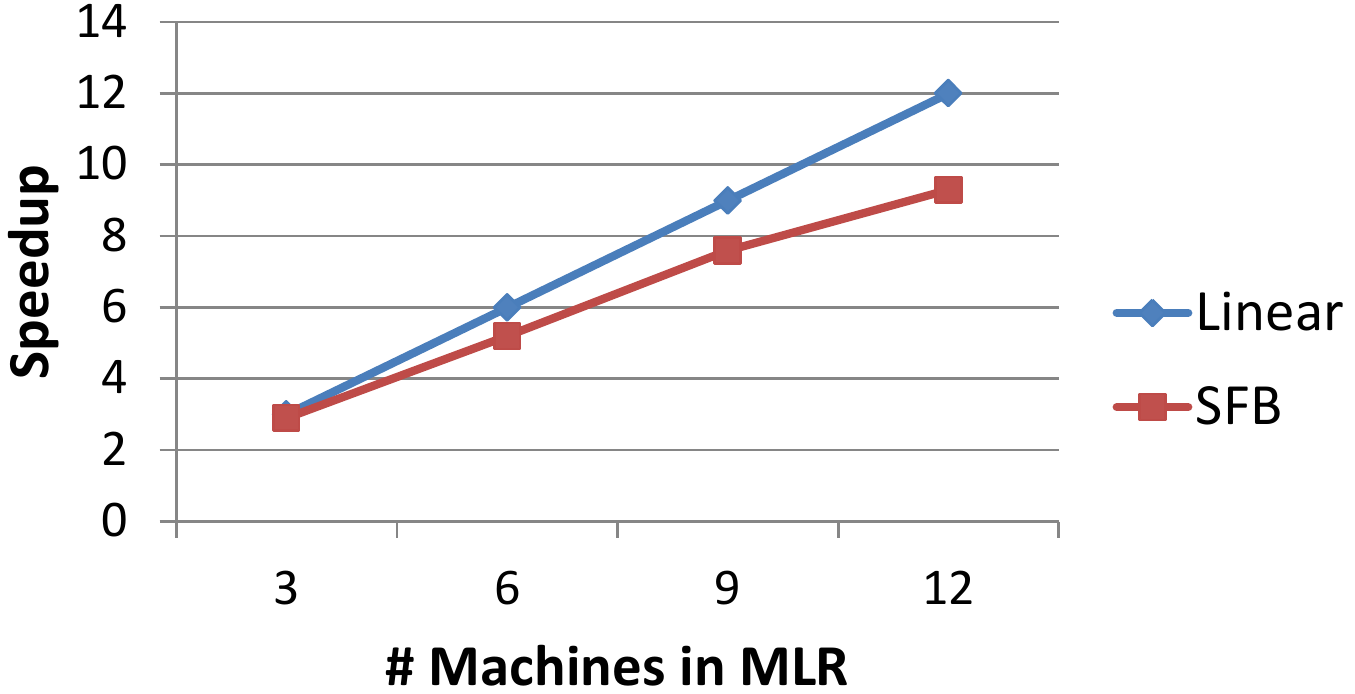}
\includegraphics[width=0.24\columnwidth]{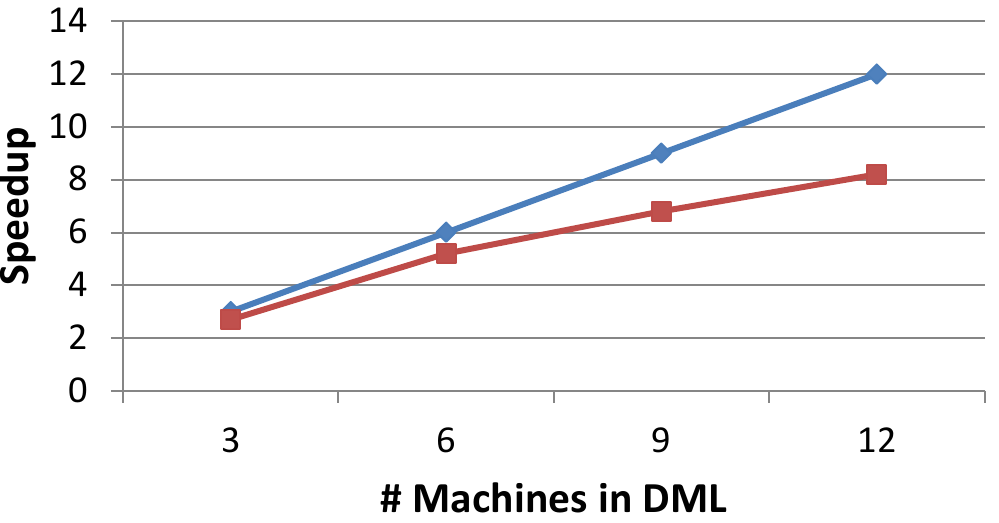}
\includegraphics[width=0.24\columnwidth]{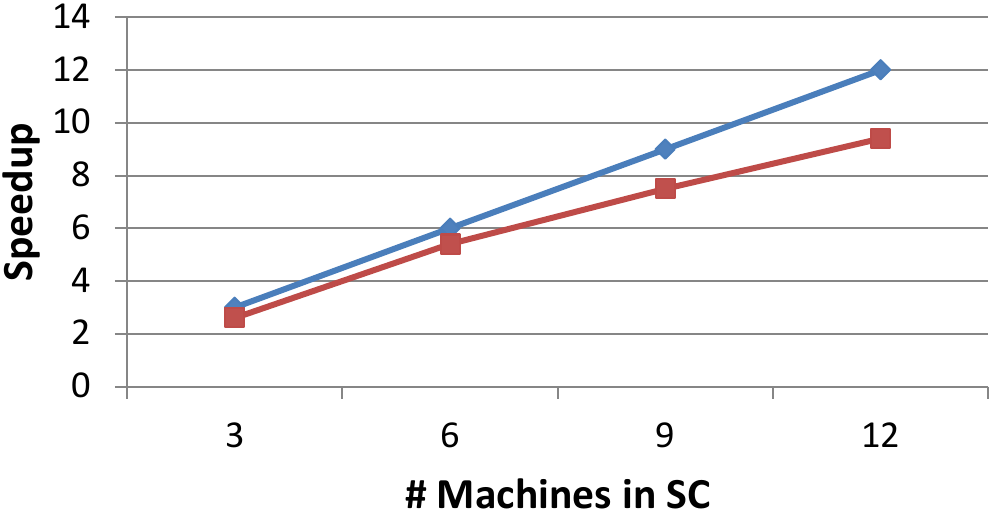}
\includegraphics[width=0.24\columnwidth]{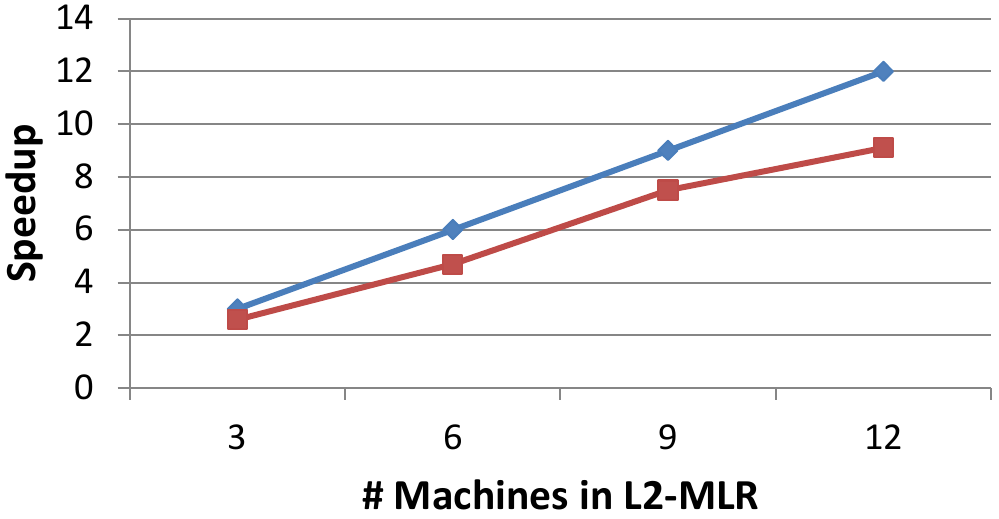}
\caption{SFB scalability with varying machines under BSP, for MLR, DML, SC, L2-MLR (left to right).}
\label{fig:exp_scalability}
\end{center}
\vspace{-0.2in}
\end{figure}

\begin{figure}[t]
\begin{center}
\includegraphics[width=0.24\columnwidth]{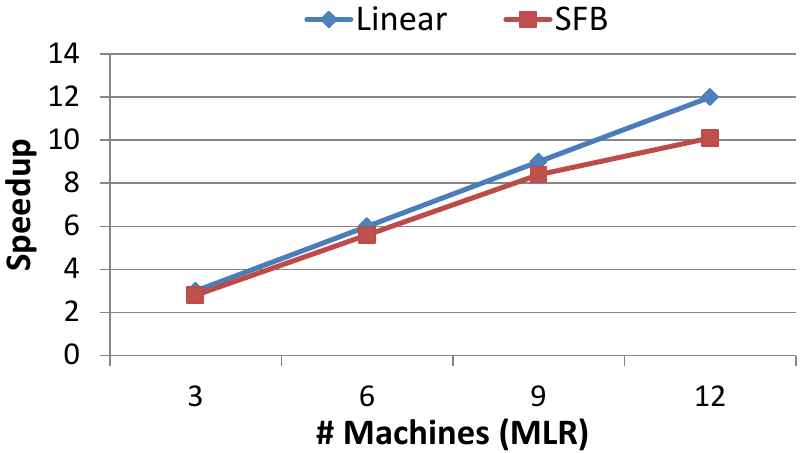}
\includegraphics[width=0.24\columnwidth]{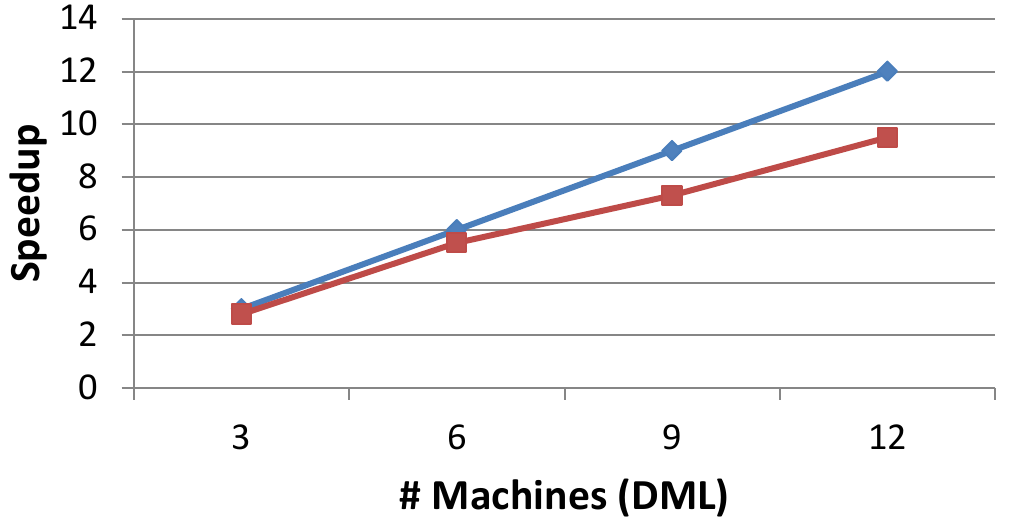}
\includegraphics[width=0.24\columnwidth]{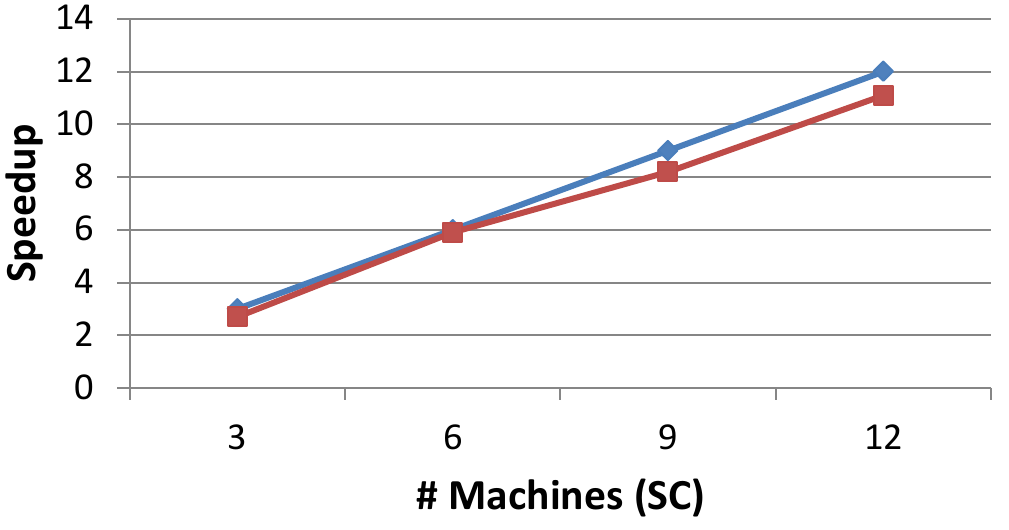}
\includegraphics[width=0.24\columnwidth]{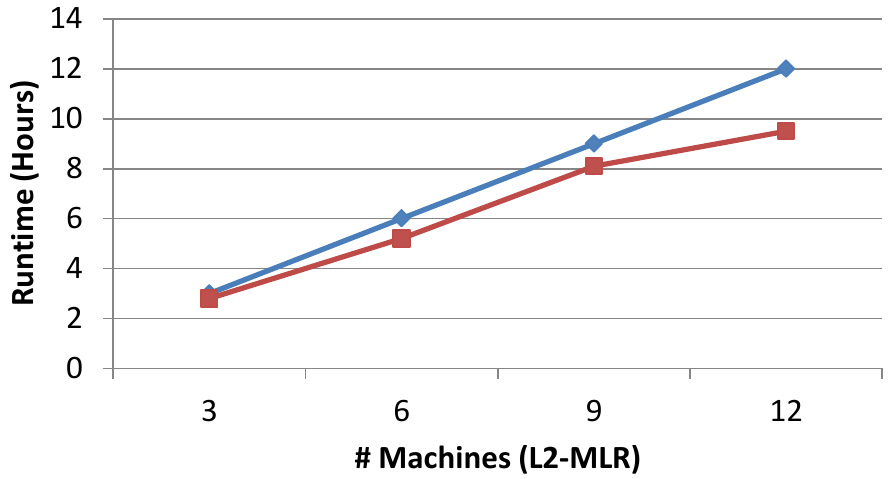}
\vspace{-0.1in}
\caption{SFB scalability with varying machines, for MLR, DML, SC, L2-MLR (left to right), under SSP (staleness=20).}
\label{fig:exp_scalability_ssp}
\end{center}
\vspace{-0.1in}
\end{figure}

\begin{figure}[t]
\begin{center}
\includegraphics[width=0.24\columnwidth]{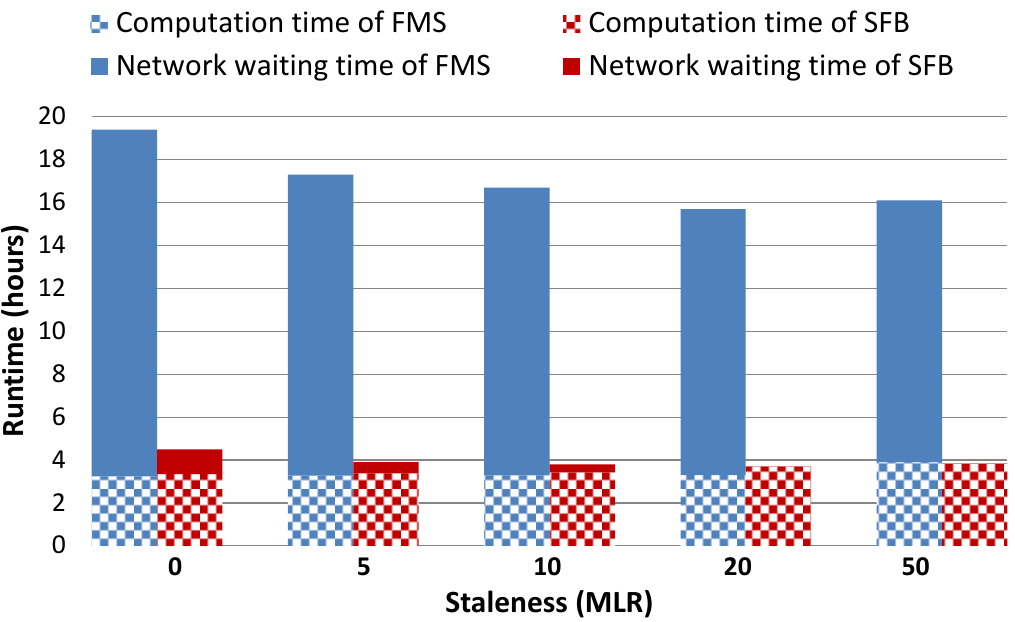}
\includegraphics[width=0.24\columnwidth]{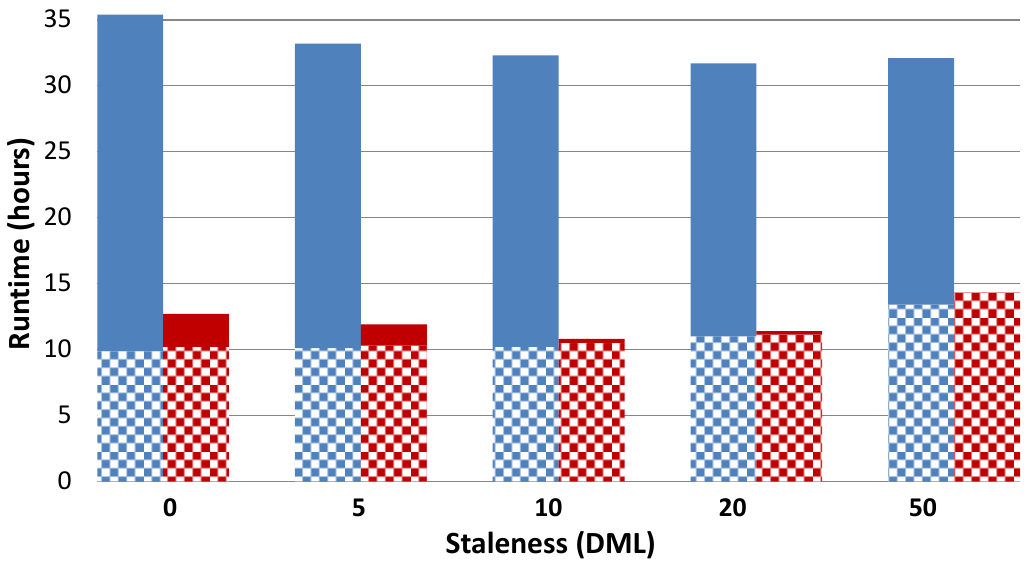}
\includegraphics[width=0.24\columnwidth]{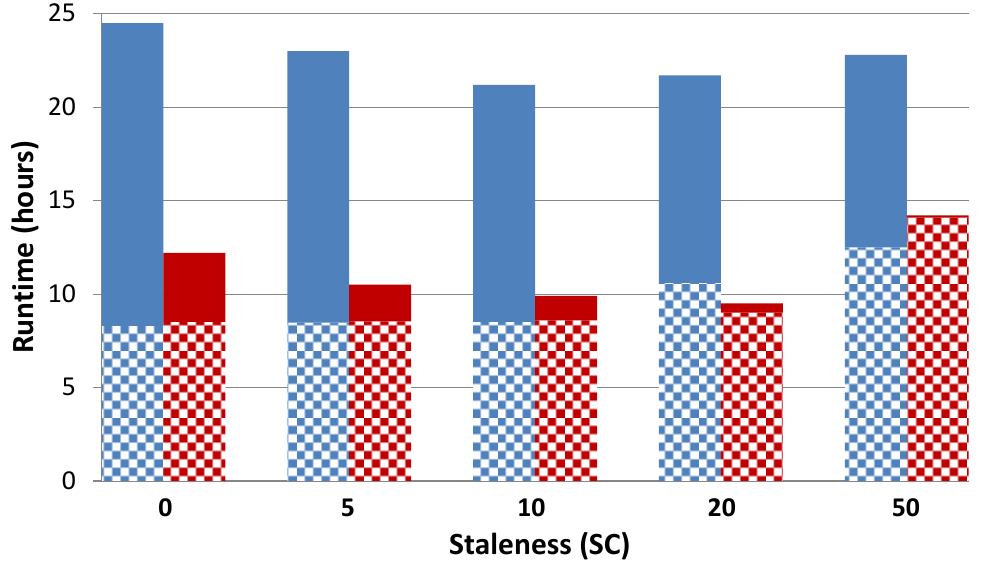}
\includegraphics[width=0.24\columnwidth]{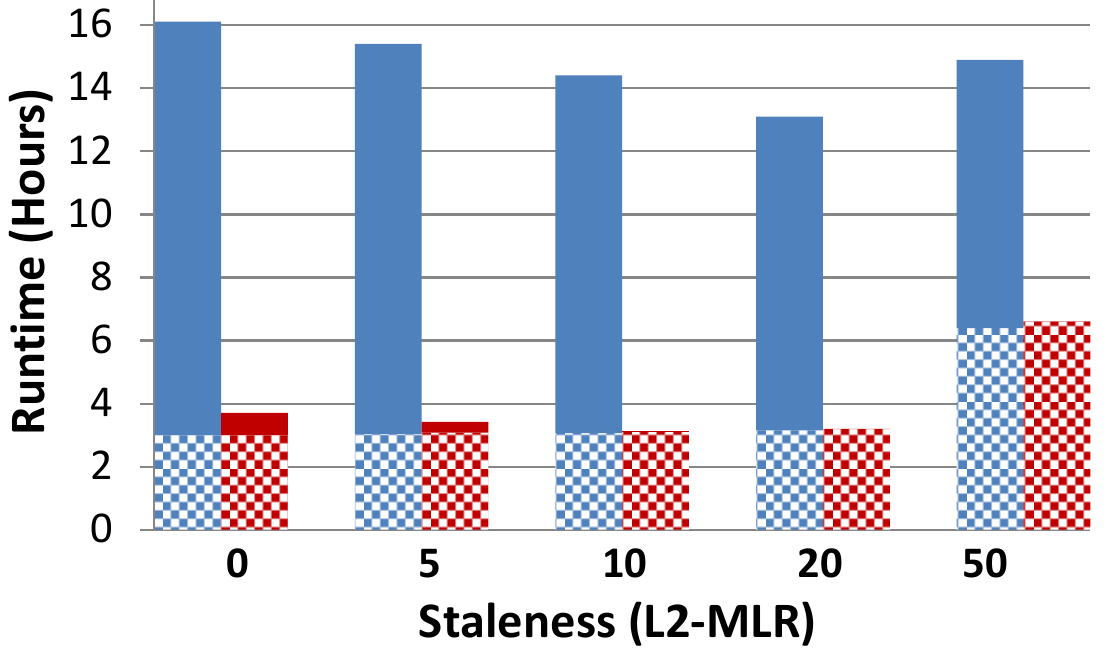}
\vspace{-0.1in}
\caption{Computation vs network waiting time for MLR, DML, SC, L2-MLR (left to right).}
\label{fig:exp_timebreak}
\end{center}
\vspace{-0.3in}
\end{figure}

We used two datasets for our experiments: (1) ImageNet \cite{deng2009imagenet} ILSFRC2012 dataset, which contains 1.2 million images from 1000 categories; the images are represented with LLC features \cite{wang2010locality}, whose dimensionality is 172k. (2) Wikipedia \cite{partalas2015lshtc} dataset, which contains 2.4 million documents from 325k categories; documents are represented with td-idf, with a dimensionality of 20k.
We ran MLR, DML, SC, L2-MLR on the Wikipedia, ImageNet, ImageNet, Wikipedia datasets respectively, and the parameter matrices contained up to 6.5b, 8.6b, 8.6b, 6.5b entries respectively (the largest latent dimension for DML and largest dictionary size for SC were both 50k).
The tradeoff parameters in SC and L2-MLR were set to 0.001 and 0.1.
We tuned the minibatch size, and found that $K=100$ was near-ideal for all experiments.
All experiments used the same constant learning rate (tuned in the range $[10^{-5},1]$). 

\vspace{-0.1in}
\paragraph{Convergence Speed and Quality}

Figure \ref{fig:exp_runtime} shows the time taken to reach a fixed objective value, for different model sizes, using BSP consistency. Figure \ref{fig:exp_runtime_ssp} shows the convergence time versus model size for MLR, DML, SC, L2-MLR, under SSP with staleness=20.
SFB converges faster than FMS, as well as Spark v1.3.1\footnote{Spark is about 2x slower than PS \cite{ho2013more,li2014scaling} based C++ implementation of FMS, due to JVM and RDD overheads.}. This is because SFB has lower communication costs, hence a greater proportion of running time gets spent on computation rather than network waiting.
This is shown in Figure \ref{fig:iter_qtt_qlt}, which plots data samples processed per second\footnote{We use samples per second instead of iterations, so different minibatch sizes can be compared.} (throughput) and algorithm progress per sample for MLR, under BSP consistency and varying minibatch sizes. The middle graph shows that SFB processes far more samples per second than FMS,
while the rightmost graph shows that SFB and FMS produce exactly the same algorithm progress per sample under BSP. For this experiment, minibatch sizes between $K=10$ and 100 performed the best as indicated by the leftmost graph.
We point out that larger model sizes should further improve SFB's advantage over FMS, because SFB has linear communications cost in the matrix dimensions, whereas FMS has quadratic costs.

\vspace{-0.1in}
\paragraph{Scalability}
In all experiments that follow, we set the number of (L2)-MLR classes, DML latent dimension, SC dictionary size to 325k, 50k, 50k respectively.
Figure \ref{fig:exp_scalability} shows SFB scalability with varying machines under BSP, for MLR, DML, SC, L2-MLR.
Figure \ref{fig:exp_scalability_ssp} shows how SFB scales with machine count, under SSP with staleness=20. 
In general, we observed close to linear (ideal) speedup, with a slight drop at 12 machines. Future work will focus on reducing peer-to-peer broadcast costs (e.g. via Halton sequences), in order to maintain linear scalability with more machines. 

\vspace{-0.1in}
\paragraph{Computation Time vs Network Waiting Time}
Figure \ref{fig:exp_timebreak} shows the {\it total} computation and network time required for SFB and FMS to converge, across a range of SSP staleness values\footnote{The Spark implementation does not easily permit this time breakdown, so we omit it.} --- in general,
higher communication cost and lower staleness induce more network waiting.
For all staleness values, SFB requires far less network waiting (because SFs are much smaller than full matrices in FMS).
Computation time for SFB is slightly longer than FMS because (1) update matrices must be reconstructed on each SFB worker, and (2) SFB requires a few more iterations for convergence, because peer-to-peer communication causes a slightly more parameter inconsistency under staleness.
Overall, the SFB reduction in network waiting time remains far greater than the added computation time, and outperforms FMS in total time. For both FMS and SFB, the shortest convergence times are achieved at moderate staleness values, confirming the importance of bounded-asynchronous communication.
\vspace{-0.1in}
\section{Related Works and Discussion}
\vspace{-0.1in}
A number of system and algorithmic solutions have been proposed to reduce communication cost in distributed ML. On the system side, 
\cite{dean2012large} proposed to reduce communication overhead by reducing the frequency of parameter/gradient exchanges between workers and the central server. 
\cite{li2014scaling} used filters to select part of ``important'' parameters/updates for transmission to reduce the number of data entries to be communicated. 
On the algorithm side, \cite{tsianos2012communication} and \cite{yang2013trading} studied the tradeoffs between communication and computation in distributed dual averaging and distributed stochastic dual coordinate ascent respectively.  \cite{shamir2013communication} proposed an approximate Newton-type method to achieve communication efficiency in distributed optimization. SFB is orthogonal to these existing approaches and be potentially combined with them to further reduce communication cost.

Peer-to-peer, decentralized architectures have been investigated in other distributed ML frameworks \cite{bhaduri2008distributed,das2010local,ormandi2013gossip,li2015malt}. Our SFBcaster system also adopt such an architecture, but with the specific purpose of supporting the SFB computation model, which is not explored by existing peer-to-peer ML frameworks.

For very large models, the size of the local parameter matrix $\mb{W}$ may exceed each machine's memory capacity --- to address this issue, we would like to investigate partitioning $\mb{W}$ over a small number of nearby machines, or using out-of-core (disk-based) storage to hold $\mb{W}$ in future work.

Finally, a promising extension to the SF idea is to re-parameterize the model $\mb{W}$ completely in terms of SFs, rather than just the updates. For example, if we initialize the parameter matrix $\mb{W}$ to be of low rank $R$, i.e., $\mb{W}^0=\sum_{j=1}^{R}\mb{u}_j\mb{v}_j^\top$, after $I$ iterations (updates), $\mb{W}^{I}=\sum_{j=1}^{R+I}\mb{u}_j\mb{v}_j^\top$. Leveraging this fact, for SGD without proximal operation and SDCA where $h(\cdot)$ is a $\ell_2$ regularizer, we can re-parametrize $\mb{W}^{I}$ using a set of SFs $\{(\mb{u}_j,\mb{v}_j)\}_{j=1}^{R+I}$, rather than maintaining $\mb{W}$ explicitly. This re-parametrization can possibly reduce both computation and storage cost, which we will investigate in the future. 

\begin{appendix}
\section{Proof of Convergence}

\subsection*{Proof of Theorem 1:}
\begin{proof}

\newcommand{\EE}{\mathbb{E}}
\newcommand{\Fcal}{\mathcal{F}}

Let $\Fcal^c := \sigma\{I_p^\tau: p=1,\ldots, P, \tau = 1, \ldots, c \}$ be the filtration generated by the random samplings $I_{p}^\tau$ up to iteration counter $c$, \ie, the information up to iteration $c$. Note that for all $p$ and $c$, $\mb{W}_p^c$ and $\mb{W}^c$ are $\Fcal^{c-1}$ measurable (since $\tau_p^q(c) \leq c-1$ by assumption), and $I_p^c$ is independent of $\Fcal^{c-1}$. Recall that the partial update generated by machine $p$ at its $c$-th iteration is
$$U_p(\mb{W}_p^c, I_p^c) = -\eta_c |S_p| \sum_{j\in I_p^c}\nabla f_{j}(\mathbf{W}_p^c)$$
Then it holds that 
$$U_p(\mathbf{W}_p^c) = \EE[ U_p(\mb{W}_p^c, I_p^c)  | \Fcal^{c-1}] = -\eta_c\nabla F_p(\mathbf{W}_p^c).$$ 
(Note that we have suppressed the dependence of $U_p$ on the iteration counter $c$.)

Then, we have
\begin{flalign}
\EE \left[\sum_{p=1}^{P} U_p(\mathbf{W}_p^c, I_p^c) ~|~ \Fcal^{c-1} \right] = \sum_{p=1}^{P} \EE [ U_p(\mathbf{W}_p^c, I_p^c) ~|~ \Fcal^{c-1} ] = \sum_{p=1}^{P} U_p(\mathbf{W}_p^c). 
\end{flalign}
Similarly we have
\begin{flalign}
\EE\left[\big\|\sum_{p=1}^{P} U_p(\mathbf{W}_p^c, I_p^c)\big\|_2^2 ~|~ \Fcal^{c-1} \right] &= \sum_{p,q=1}^{P}\EE[\langle U_p(\mathbf{W}_p^c, I_p^c), U_q(\mathbf{W}_q^c, I_q^c) \rangle ~|~ \Fcal^{c-1} ] \\
&= \sum_{p,q = 1}^{P} \langle U_p(\mathbf{W}_q^c), U_q(\mathbf{W}_q^c) \rangle \\
&\quad+ \sum_{p=1}^{P} \EE\left[ \|U_p(\mathbf{W}_p^c, I_p^c) - U_p(\mathbf{W}_p^c)\|_2^2 ~|~ \Fcal^{c-1} \right].
\end{flalign}
The variance term in the above equality can be bounded as
\begin{flalign*}
\sum_{p=1}^{P} \mathbb{E}\left[\|U_p(\mathbf{W}_p^c, I_p^c) - U_p(\mathbf{W}_p^c)\|_2^2 ~|~ \Fcal^{c-1} \right] &= \eta_c^2 \underbrace{\sum_{p=1}^{P} \mathbb{E}\left[\||S_p|\sum_{j\in I_p^c}\nabla f_j(\mathbf{W}_p^c) - \nabla F_p(\mathbf{W}_p^c)\|_2^2 ~|~ \Fcal^{c-1} \right]}_{\hat{\sigma}^2 P}\\
&\le  \eta_c^2 \hat{\sigma}^2 P,
\end{flalign*}
Now use the update rule $\mathbf{W}_p^{c+1}  = \mathbf{W}_p^c + \sum_{p=1}^{P}U_p(\mathbf{W}_p^c, I_p^c)$ and  the descent lemma \cite{BertsekasTsitsiklis89}, we have
\begin{flalign}
F(\mathbf{W}^{c+1}) -F(\mathbf{W}^c) &\le \langle \mathbf{W}^{c+1} - \mathbf{W}^c, \nabla F(\mathbf{W}^c) \rangle + \frac{L_F}{2}\|\mathbf{W}^{c+1} - \mathbf{W}^c\|_2^2 \label{eq: 1}\\
&= \langle\sum_{p=1}^{P}U_p(\mathbf{W}_p^c, I_p^c), \nabla F(\mathbf{W}^c) \rangle + \frac{L_F}{2}\|\sum_{p=1}^{P}U_p(\mathbf{W}_p^c, I_p^c)\|_2^2 \label{eq: 2}
\end{flalign}
Then take expectation on both sides, we obtain
\begin{flalign}
\mathbb{E}\left[\right. & \left.F(\mathbf{W}^{c+1})  -F(\mathbf{W}^c)~|~ \Fcal^{c-1}\right] \le \langle \sum_{p=1}^{P}U_p(\mathbf{W}_p^c), \nabla F(\mathbf{W}^c)\rangle  + \frac{L_F}{2}\|\sum_{p=1}^{P}U_p(\mathbf{W}_p^c)\|_2^2+ \frac{L_F \eta_c^2 \hat\sigma^2 P}{2} \label{eq: 3}\\
&= (\frac{L_F}{2} - \eta_c^{-1})\|\sum_{p=1}^{P}U_p(\mathbf{W}_p^c)\|_2^2 - \eta_c^{-1}\langle \sum_{p=1}^{P}U_p(\mathbf{W}_p^c), \sum_{p=1}^{P}[U_p(\mathbf{W}^c)- U_p(\mathbf{W}_p^c)]\rangle + \frac{L_F \eta_c^2 \hat\sigma^2 P}{2}\label{eq: 5}\\
&\le (\frac{L_F}{2} - \eta_c^{-1})\|\sum_{p=1}^{P}U_p(\mathbf{W}_p^c)\|_2^2 + \|\sum_{p=1}^{P}U_p(\mathbf{W}_p^c)\|\sum_{p=1}^{P} L_p \|\mathbf{W}^c - \mathbf{W}_p^c\| +\frac{L_F \eta_c^2 \hat\sigma^2 P}{2}\label{eq: 6},
\end{flalign}
Now take expectation w.r.t all random variables, we obtain
\begin{flalign}
\EE\left[F(\mathbf{W}^{c+1}) -F(\mathbf{W}^c)\right] &\le (\frac{L_F}{2} - \eta_c^{-1})\EE\left[\|\sum_{p=1}^{P}U_p(\mathbf{W}_p^c)\|_2^2\right] \\
\label{eq:tmp}
&\quad + \sum_{p=1}^{P} L_p\EE\left[\|\sum_{p=1}^{P}U_p(\mathbf{W}_p^c)\| \|\mathbf{W}^c - \mathbf{W}_p^c\|\right] +\frac{L_F \eta_c^2 \sigma^2 P}{2}
\end{flalign}

Next we proceed to bound the term $\mathbb{E}\|\sum_{p=1}^{P}U_p(\mathbf{W}_p^c)\| \|\mathbf{W}^c - \mathbf{W}_p^c\|$. We list the auxiliary update rule and the local update rule here for convenience.
\begin{flalign}
\mathbf{W}^c &= \mathbf{W}^0 + \sum_{q=1}^{P}\sum_{t=0}^{c-1}U_q(\mathbf{W}_q^t, I_q^t), \\
\mathbf{W}_p^c &= \mathbf{W}^0 +  \sum_{q=1}^{P}\sum_{t=0}^{\tau_p^q(c)} U_q(\mathbf{W}_q^t, I_q^t).
\end{flalign}
Now subtract the above two and use the bounded delay assumption $0\le (c-1) - \tau_p^q(c) \le s$, we obtain
\begin{flalign}
\|\mathbf{W}^c - \mathbf{W}_p^c\| &= \|\sum_{q=1}^{P}\sum_{t=\tau_p^q(c)+1}^{c-1}  U_q(\mathbf{W}_q^t, I_q^t) \|\\
&\le \|\sum_{q=1}^{P}\sum_{t=c-s}^{c-1}  U_q(\mathbf{W}_q^t, I_q^t) \|+ \|\sum_{q=1}^{P} \sum_{t=c-s}^{\tau_p^q(c)} U_q(\mathbf{W}_q^t, I_q^t)\|\\
\label{eq:tmp2}
&\le \sum_{t=c-s}^{c-1} \|\sum_{q=1}^{P}U_q(\mathbf{W}_q^t, I_q^t)\| + \eta_{c-s} G,
\end{flalign}
where the last inequality follows from the facts that $\eta_c$ is strictly decreasing, and $\|\sum_{q=1}^{P} \sum_{t=c-s}^{\tau_p^q(c)} \nabla F_q(\mathbf{W}_q^t, I_q^t)\|$ is bounded by some constant $G$ since $\nabla F_q$ is continuous and all the sequences $\mathbf{W}_p^c$ are bounded. Thus by taking expectation, we obtain
\begin{flalign}
\EE\left[\|\sum_{p=1}^{P}U_p(\mathbf{W}_p^c)\| \right. & \left. \|\mathbf{W}^c - \mathbf{W}_p^c\|\right] \le \EE \left[\|\sum_{p=1}^{P}U_p(\mathbf{W}_p^c)\|\Big(\sum_{t=c-s}^{c-1} \|\sum_{q=1}^{P}U_q(\mathbf{W}_q^t, I_q^t)\|+\eta_{c-s} G\Big) \right]\\
&=\sum_{t=c-s}^{c-1}\EE\left[\|\sum_{p=1}^{P}U_p(\mathbf{W}_p^c)\| \|\sum_{q=1}^{P}U_q(\mathbf{W}_q^t, I_q^t)\|\right] + \eta_{c-s} G \cdot \EE\left[\|\sum_{p=1}^{P}U_p(\mathbf{W}_p^c)\|\right] \\
&\le \sum_{t=c-s}^{c-1}\EE\left[\|\sum_{p=1}^{P}U_p(\mathbf{W}_p^c)\|_2^2+ \|\sum_{q=1}^{P}U_q(\mathbf{W}_q^t, I_q^t)\|_2^2\right] + \EE\|\sum_{p=1}^{P}U_p(\mathbf{W}_p^c)\|_2^2 + \eta_{c-s}^2 G^2\\
&\le (s+1)\EE\|\sum_{p=1}^{P}U_p(\mathbf{W}_p^c)\|_2^2 + \sum_{t=c-s}^{c-1}\left[\EE\|\sum_{q=1}^{P}U_q(\mathbf{W}_q^t)\|_2^2 +\eta_t^2 \sigma^2 P\right] + \eta_{c-s}^2 G^2.
\end{flalign}
Now plug this into the previous result in \eqref{eq:tmp}:
\begin{flalign}
\EE F(\mathbf{W}^{c+1}) -\EE F(\mathbf{W}^c) &\le (\frac{L_F}{2} - \eta_c^{-1})\EE \|\sum_{p=1}^{P}U_p(\mathbf{W}_p^c)\|_2^2 +(s+1)L\EE \|\sum_{p=1}^{P}U_p(\mathbf{W}_p^c)\|_2^2 \\
&\quad+ \sum_{t=c-s}^{c-1}\left[L\EE \|\sum_{p=1}^{P}U_p(\mathbf{W}_p^c)\|_2^2 + \eta_t^2 L\sigma^2 P\right] + \eta_{c-s}^2 G^2 L + \frac{L_F \eta_c^2 \sigma^2 P}{2}\\
&=(\frac{L_F}{2}+(s+1)L - \eta_c^{-1})\EE \|\sum_{p=1}^{P}U_p(\mathbf{W}_p^c)\|_2^2\\
&\quad+ \sum_{t=c-s}^{c-1}\left[L\EE \|\sum_{p=1}^{P}U_p(\mathbf{W}_p^c)\|_2^2 + \eta_t^2 L\sigma^2 P\right] +\eta_{c-s}^2 G^2 L+\frac{L_F \eta_c^2 \sigma^2 P}{2}
\end{flalign}
Sum both sides over $c = 0,...,C$:
\begin{flalign}
\EE F(\mathbf{W}^{C+1}) -\EE F(\mathbf{W}^0) &\le\sum_{c=0}^{C}\left[(\frac{L_F}{2}+(2s+1)L - \eta_c^{-1})\EE \|\sum_{p=1}^{P}U_p(\mathbf{W}_p^c)\|_2^2\right]\\
&\quad+(L\sigma^2 P s+\frac{L_F \sigma^2 P}{2}) \sum_{c=0}^{C}\eta_c^2 + G^2 L\sum_{c=0}^{C}\eta_{c-s}^2.
\end{flalign}
After rearranging terms we finally obtain
\begin{flalign}
\sum_{c=0}^{C}\left[\eta_c^2 (\eta_c^{-1} - \frac{L_F}{2}-2(s+1)L)\EE\|\sum_{p=1}^{P} \right.
& \left. \nabla F_p(\mathbf{W}_p^c)\|_2^2\right]\le  \EE F(\mathbf{W}^0) - \EE F(\mathbf{W}^{C+1}) \\
&\quad+ (L\sigma^2 P s+\frac{L_F \sigma^2 P}{2}) \sum_{c=0}^{C}\eta_c^2 + G^2 L\sum_{c=0}^{C}\eta_{c-s}^2.
\end{flalign}

Now set $\eta_c^{-1} = \frac{L_F}{2}+2sL +\sqrt{c}$. Then, the above inequality becomes (ignoring some universal constants):
\begin{flalign}
\label{eq:rate}
\sum_{c=0}^{C}\left[\frac{1}{\sqrt{c}}\mathbb{E}\|\sum_{p=1}^{P}\nabla F_p(\mathbf{W}_p^c)\|_2^2\right] \le O\left(\Big((L+L_F)\sigma^2 P s\Big)\sum_{c=0}^{C}\frac{1}{c}\right).
\end{flalign}
Since $\sum_{c=0}^{C}\frac{1}{c} = o(\sum_{c=0}^{C}\frac{1}{\sqrt{c}})$, we must have 
\begin{align}
\label{eq:claim1}
\liminf\limits_{c\to\infty}\mathbb{E}\|\sum_{p=1}^{P}\nabla F_p(\mathbf{W}_p^c)\| = 0,
\end{align} 
proving the first claim.

On the other hand, the bound of $\|\mathbf{W}^c - \mathbf{W}_p^c\|$ in \eqref{eq:tmp2} gives
\begin{flalign}
\|\mathbf{W}^c - \mathbf{W}_p^c\| \le \sum_{t=c-s}^{c-1} \eta_t \|\sum_{q=1}^{P} |S_q| \sum_{j\in I_q^t}\nabla f_{j}(\mathbf{W}_q^t)\| + \eta_{c-s} G.
\end{flalign}
By assumption the sequences $\{\mb{W}_p^c\}_{p,c}$ and $\{\mb{W}^c\}_c$ are bounded and the gradient of $f_j$ is continuous, thus $\nabla f_{j}(\mathbf{W}_q^t)$ is bounded. Now take $c\rightarrow \infty$ in the above inequality and notice that $\underset{c\rightarrow \infty}{\mathrm{lim}}\eta_c = 0$, we have $\underset{c\rightarrow \infty}{\mathrm{lim}}\|\mathbf{W}^c - \mathbf{W}_p^c\| = 0$ almost surely, proving the second claim.

Lastly, the Lipschitz continuity of $\nabla F_p$ further implies
$$0 = \liminf\limits_{c\to\infty}\mathbb{E}\|\sum_{p=1}^{P}\nabla F_p(\mathbf{W}_p^c)\| \geq \liminf\limits_{c\to\infty}\mathbb{E}\|\sum_{p=1}^{P}\nabla F_p(\mathbf{W}^c)\| = \liminf\limits_{c\to\infty}\mathbb{E}\|\nabla F(\mathbf{W}^c)\|=0.$$
Thus there exists a common limit point of $\mathbf{W}^c, \mathbf{W}_p^c$ that is a stationary point almost surely. From \eqref{eq:rate} and use the estimate $\sum_{c=1}^C \tfrac{1}{c} \approx \log C$, we have 
\begin{align}
\min_{c=1,\ldots, C} \EE \left[\|\sum_{p=1}^{P}\nabla F_p(\mathbf{W}_p^c)\|_2^2\right] \leq O\left(\frac{(L+L_F)\sigma^2 P s\log C}{\sqrt{C}}\right).
\end{align}
The proof is now complete.
\end{proof}



\end{appendix}

{\small
\bibliographystyle{abbrv}
\bibliography{svb}

\begin{thebibliography}{10}

\bibitem{agarwal2011distributed}
A.~Agarwal and J.~C. Duchi.
\newblock Distributed delayed stochastic optimization.
\newblock In {\em NIPS}, 2011.

\bibitem{ahmed2012scalable}
A.~Ahmed, M.~Aly, J.~Gonzalez, S.~Narayanamurthy, and A.~J. Smola.
\newblock Scalable inference in latent variable models.
\newblock In {\em WSDM}, 2012.

\bibitem{beck2009fast}
A.~Beck and M.~Teboulle.
\newblock A fast iterative shrinkage-thresholding algorithm for linear inverse
  problems.
\newblock {\em SIAM Journal on Imaging Sciences}, 2009.

\bibitem{bertsekas1999nonlinear}
D.~P. Bertsekas.
\newblock {\em Nonlinear programming}.
\newblock Athena scientific Belmont, 1999.

\bibitem{BertsekasTsitsiklis89}
D.~P. Bertsekas and J.~N. Tsitsiklis.
\newblock {\em Parallel and Distributed Computation: Numerical Methods}.
\newblock Prentice-Hall, 1989.

\bibitem{bhaduri2008distributed}
K.~Bhaduri, R.~Wolff, C.~Giannella, and H.~Kargupta.
\newblock Distributed decision-tree induction in peer-to-peer systems.
\newblock {\em Statistical Analysis and Data Mining: The ASA Data Science
  Journal}, 2008.

\bibitem{chilimbi2014project}
T.~Chilimbi, Y.~Suzue, J.~Apacible, and K.~Kalyanaraman.
\newblock Project adam: building an efficient and scalable deep learning
  training system.
\newblock In {\em OSDI}, 2014.

\bibitem{dai2015high}
W.~Dai, A.~Kumar, J.~Wei, Q.~Ho, G.~Gibson, and E.~P. Xing.
\newblock High-performance distributed ml at scale through parameter server
  consistency models.
\newblock In {\em AAAI}. 2015.

\bibitem{das2010local}
K.~Das, K.~Bhaduri, and H.~Kargupta.
\newblock A local asynchronous distributed privacy preserving feature selection
  algorithm for large peer-to-peer networks.
\newblock {\em Knowledge and information systems}, 2010.

\bibitem{dean2012large}
J.~Dean, G.~Corrado, R.~Monga, K.~Chen, M.~Devin, M.~Mao, A.~Senior, P.~Tucker,
  K.~Yang, Q.~V. Le, et~al.
\newblock Large scale distributed deep networks.
\newblock In {\em NIPS}, 2012.

\bibitem{dean2008mapreduce}
J.~Dean and S.~Ghemawat.
\newblock Mapreduce: simplified data processing on large clusters.
\newblock {\em CACM}, 2008.

\bibitem{deng2009imagenet}
J.~Deng, W.~Dong, R.~Socher, L.-J. Li, K.~Li, and L.~Fei-Fei.
\newblock Imagenet: A large-scale hierarchical image database.
\newblock In {\em CVPR}, 2009.

\bibitem{gonzalez2012powergraph}
J.~E. Gonzalez, Y.~Low, H.~Gu, D.~Bickson, and C.~Guestrin.
\newblock Powergraph: distributed graph-parallel computation on natural graphs.
\newblock In {\em OSDI}, 2012.

\bibitem{gopal2013distributed}
S.~Gopal and Y.~Yang.
\newblock Distributed training of large-scale logistic models.
\newblock In {\em ICML}, 2013.

\bibitem{ho2013more}
Q.~Ho, J.~Cipar, H.~Cui, S.~Lee, J.~K. Kim, P.~B. Gibbons, G.~A. Gibson,
  G.~Ganger, and E.~Xing.
\newblock More effective distributed ml via a stale synchronous parallel
  parameter server.
\newblock In {\em NIPS}, 2013.

\bibitem{hsieh2008dual}
C.-J. Hsieh, K.-W. Chang, C.-J. Lin, S.~S. Keerthi, and S.~Sundararajan.
\newblock A dual coordinate descent method for large-scale linear svm.
\newblock In {\em ICML}, 2008.

\bibitem{hsieh2015comm}
C.-J. Hsieh, H.-F. Yu, and I.~S. Dhillon.
\newblock Passcode: Parallel asynchronous stochastic dual co-ordinate descent.
\newblock In {\em ICML}, 2015.

\bibitem{jaggi2014communication}
M.~Jaggi, V.~Smith, M.~Tak{\'a}c, J.~Terhorst, S.~Krishnan, T.~Hofmann, and
  M.~I. Jordan.
\newblock Communication-efficient distributed dual coordinate ascent.
\newblock In {\em NIPS}, 2014.

\bibitem{lee1999learning}
D.~D. Lee and H.~S. Seung.
\newblock Learning the parts of objects by non-negative matrix factorization.
\newblock {\em Nature}, 1999.

\bibitem{lee2014strads}
S.~Lee, J.~K. Kim, X.~Zheng, Q.~Ho, G.~A. Gibson, and E.~P. Xing.
\newblock On model parallelization and scheduling strategies for distributed
  machine learning.
\newblock {\em NIPS}, 2014.

\bibitem{li2015malt}
H.~Li, A.~Kadav, E.~Kruus, and C.~Ungureanu.
\newblock Malt: distributed data-parallelism for existing ml applications.
\newblock In {\em Proceedings of the Tenth European Conference on Computer
  Systems}, 2015.

\bibitem{li2014scaling}
M.~Li, D.~G. Andersen, J.~W. Park, A.~J. Smola, A.~Ahmed, V.~Josifovski,
  J.~Long, E.~J. Shekita, and B.-Y. Su.
\newblock Scaling distributed machine learning with the parameter server.
\newblock In {\em OSDI}, 2014.

\bibitem{malewicz2010pregel}
G.~Malewicz, M.~H. Austern, A.~J. Bik, J.~C. Dehnert, I.~Horn, N.~Leiser, and
  G.~Czajkowski.
\newblock Pregel: a system for large-scale graph processing.
\newblock In {\em SIGMOD}, 2010.

\bibitem{olshausen1997sparse}
B.~A. Olshausen and D.~J. Field.
\newblock Sparse coding with an overcomplete basis set: A strategy employed by
  v1?
\newblock {\em Vision research}, 1997.

\bibitem{ormandi2013gossip}
R.~Orm{\'a}ndi, I.~Heged{\H{u}}s, and M.~Jelasity.
\newblock Gossip learning with linear models on fully distributed data.
\newblock {\em Concurrency and Computation: Practice and Experience}, 2013.

\bibitem{partalas2015lshtc}
I.~Partalas, A.~Kosmopoulos, N.~Baskiotis, T.~Artieres, G.~Paliouras,
  E.~Gaussier, I.~Androutsopoulos, M.-R. Amini, and P.~Galinari.
\newblock Lshtc: A benchmark for large-scale text classification.
\newblock {\em arXiv:1503.08581 [cs.IR]}, 2015.

\bibitem{shalev2013stochastic}
S.~Shalev-Shwartz and T.~Zhang.
\newblock Stochastic dual coordinate ascent methods for regularized loss.
\newblock {\em JMLR}, 2013.

\bibitem{shamir2013communication}
O.~Shamir, N.~Srebro, and T.~Zhang.
\newblock Communication efficient distributed optimization using an approximate
  newton-type method.
\newblock {\em ICML}, 2014.

\bibitem{sindhwani2012large}
V.~Sindhwani and A.~Ghoting.
\newblock Large-scale distributed non-negative sparse coding and sparse
  dictionary learning.
\newblock In {\em KDD}, 2012.

\bibitem{smolensky1986information}
P.~Smolensky.
\newblock Information processing in dynamical systems: Foundations of harmony
  theory.
\newblock 1986.

\bibitem{terry2013replicated}
D.~Terry.
\newblock Replicated data consistency explained through baseball.
\newblock {\em CACM}, 2013.

\bibitem{tsianos2012communication}
K.~Tsianos, S.~Lawlor, and M.~G. Rabbat.
\newblock Communication/computation tradeoffs in consensus-based distributed
  optimization.
\newblock In {\em NIPS}, 2012.

\bibitem{wang2010locality}
J.~Wang, J.~Yang, K.~Yu, F.~Lv, T.~Huang, and Y.~Gong.
\newblock Locality-constrained linear coding for image classification.
\newblock In {\em CVPR}, 2010.

\bibitem{weinberger2005distance}
K.~Q. Weinberger, J.~Blitzer, and L.~K. Saul.
\newblock Distance metric learning for large margin nearest neighbor
  classification.
\newblock In {\em NIPS}, 2005.

\bibitem{xing2002distance}
E.~P. Xing, M.~I. Jordan, S.~Russell, and A.~Y. Ng.
\newblock Distance metric learning with application to clustering with
  side-information.
\newblock In {\em NIPS}, 2002.

\bibitem{yang2013trading}
T.~Yang.
\newblock Trading computation for communication: Distributed stochastic dual
  coordinate ascent.
\newblock In {\em NIPS}, 2013.

\bibitem{yuan2006model}
M.~Yuan and Y.~Lin.
\newblock Model selection and estimation in regression with grouped variables.
\newblock {\em Journal of the Royal Statistical Society: Series B (Statistical
  Methodology)}, 2006.

\bibitem{zaharia2012resilient}
M.~Zaharia, M.~Chowdhury, T.~Das, A.~Dave, J.~Ma, M.~McCauley, M.~J. Franklin,
  S.~Shenker, and I.~Stoica.
\newblock Resilient distributed datasets: A fault-tolerant abstraction for
  in-memory cluster computing.
\newblock In {\em NSDI}, 2012.

\end{thebibliography}
}
\end{document}